\documentclass[journal]{IEEEtran}
\usepackage{amsmath,amsfonts,bm}

\def\eqref#1{equation~\ref{#1}}
\def\1{\bm{1}}

\def\mS{{\bm{S}}}

\DeclareMathAlphabet{\mathsfit}{\encodingdefault}{\sfdefault}{m}{sl}
\SetMathAlphabet{\mathsfit}{bold}{\encodingdefault}{\sfdefault}{bx}{n}

\usepackage{cite}
\newcommand{\citep}{\cite}
\newcommand{\citet}{\cite}
\ifCLASSINFOpdf
\else
\fi
\usepackage{hyperref}
\usepackage[linesnumbered,ruled]{algorithm2e}
\newcommand{\nonl}{\renewcommand{\nl}{\let\nl\oldnl}}
\let\oldnl\nl% Store \nl in \oldnl

\usepackage{url}
\usepackage{xspace}
\usepackage{nccmath} % for medsize
\usepackage{multirow}
\usepackage{booktabs}
\usepackage{bm}
\usepackage{graphicx}

\usepackage{xcolor}

\begin{document}
\title{Exploring the Potential of Low-bit Training of Convolutional Neural Networks}

% \author{Michael~Shell,~\IEEEmembership{Member,~IEEE,}
%         John~Doe,~\IEEEmembership{Fellow,~OSA,}
%         and~Jane~Doe,~\IEEEmembership{Life~Fellow,~IEEE}% <-this % stops a space
% \thanks{M. Shell was with the Department
% of Electrical and Computer Engineering, Georgia Institute of Technology, Atlanta,
% GA, 30332 USA e-mail: (see http://www.michaelshell.org/contact.html).}% <-this % stops a space
% \thanks{J. Doe and J. Doe are with Anonymous University.}% <-this % stops a space
% \thanks{Manuscript received April 19, 2005; revised August 26, 2015.}}

\author{Kai~Zhong,
  Xuefei~Ning,
  Guohao~Dai,
  Zhenhua~Zhu,
  Tianchen~Zhao,
  Shulin~Zeng,
  Yu~Wang,~\IEEEmembership{Senior~Member,~IEEE,}
  and~Huazhong~Yang,~\IEEEmembership{Fellow,~IEEE}% <-this % stops a space
\thanks{All authors were with the Department
  of Electronic and Computer Engineering, Tsinghua University, Beijing, China, and Beijing National Research Center for Information Science and Technology (BNRist) (e-mail: yu-wang@tsinghua.edu.cn).}}
% %<-this % stops a space
%}


\markboth{IEEE TRANSACTIONS ON COMPUTER-AIDED DESIGN OF INTEGRATED CIRCUITS AND SYSTEMS}{}



\maketitle

\begin{abstract}

Convolutional neural networks (CNNs) have been widely used in many tasks, but training CNNs is time-consuming and energy-hungry. Using the low-bit integer format has been proved promising for speeding up and improving the energy efficiency of CNN inference, while the training phase of CNNs can hardly benefit from such a technique because of following challenges: \textit{(1) The integer data format cannot meet the requirements of the data dynamic range in training, resulting in the accuracy drop; (2) The floating-point data format keeps large dynamic range with much more exponent bits, resulting in higher accumulation power than integer one; (3) There are some specially designed data formats (e.g., with group-wise scaling) that have the potential to deal with the former two problems but the common hardware can not support them efficiently.}

To tackle all these challenges and make the training phase of CNNs benefit from the low-bit format, we propose a low-bit training framework for convolutional neural networks to pursue a better trade-off between the accuracy and energy efficiency. \textit{(1) We adopt element-wise scaling to improve the dynamic range of data representation, which greatly reduces the quantization error; (2) Group-wise scaling with hardware friendly factor format is designed to reduce the element-wise exponent bits without degrading the accuracy; (3) %The low-bit tensor convolution arithmetic and dynamic quantization procedure are described and 
We design the customized hardware unit that implement the low-bit tensor convolution arithmetic with our multi-level scaling data format.} Experiments show that our framework achieves a superior trade-off between the accuracy and the bit-width than previous low-bit training studies. 
For training a variety of models on CIFAR-10, using 1-bit mantissa and 2-bit exponent is adequate to keep the accuracy loss within $1\%$. And on larger datasets like ImageNet, using 4-bit mantissa and 2-bit exponent is adequate.
Through the energy consumption simulation of the computing units,we can estimate that training a variety of models with our framework could achieve $8.3\sim10.2\times$ and $1.9\sim2.3\times$ higher energy efficiency than single-precision and 8-bit floating-point arithmetic, respectively. %\todo{(please shorten the exp part)}

\end{abstract}


% \fi
\IEEEpeerreviewmaketitle

\section{Introduction}
\label{sec:intro} 
 %and object detection~\citep{yolo,ssd}. 

\IEEEPARstart{C}{onvolutional} neural networks (CNNs) have achieved state-of-the-art performance in many computer vision tasks~\citep{alexnet,yolo,ssd}.
However, deep CNNs are both computation and storage-intensive. The training process could consume up to hundreds of ExaFLOPs of computations and tens of GBytes of storage~\citep{vgg}, thus posing a tremendous challenge for training in resource-constrained environments. 
At present, GPU is commonly used to train CNNs and it is energy-hungry. The power of a running GPU is about 250W, and it usually takes more than 10 GPU-days to train one CNN model on large practical datasets like ImageNet~\citep{imagenet}. Therefore, reducing the energy consumption of the training process has raised interest in recent years.
%exploring alternative training platforms that is more environment-friendly 

%It makes AI applications expensive and not environment-friendly.

Reducing the precision of CNNs has drawn great attention since it can %due to its potential in
reduce both the storage and computational complexity. 
It is pointed out that 32-bit floating-point multiplication and addition units consumes about $20\sim30\times$ more power than 8-bit fixed-point ones~\citep{fixfloat}. Also, using 8-bit data format could save the energy consumption of memory access by roughly 4 times.
%\zksays{It is pointed out that the power consumption of 32-bit floating-point multiplication and addition units are about 20 to 30 times those of 8-bit fixed-point ones~\citep{fixfloat}, and the energy consumption of memory access will obviously be saved 4 times.}
Many studies~\citep{quantizetraining,pact,post,hawq} focus on amending the training process to acquire a reduced-precision model with higher inference efficiency.
However, these methods rely on tuning from a full-precision pre-trained model, which is costly, or introduce more optimization operations into training for a better inference performance, therefore, they are not suitable for efficient training.
%which already costs a lot, or even introduce more optimization operations into training for a better inference performance. %computational unit design and making it hard for high-efficiency circuit design.
% The methods that use an extremely low bit-width (e.g., 1 or 2) %struggle in big datasets, and
% cause non-negligible accuracy degradation on large datasets~\citep{dorefa,xnor}.
%Unlike the optimization for inference efficiency, there are a few studies aiming at accelerating the training process. 
% and use different format in the forward and backward process to get better accuracy. 
%Moreover, since the cost of multiplications is higher than additions~\citep{fixfloat}, the multiplications in convolution account for the major computational energy consumption of the CNN training process. 
% but their formats  are not energy efficient enough, and accumulations are still full-precision. %floating-point bit-width to 8 during the training process.

\begin{table}[bt]
    \centering
    \caption{The number of different operations in the training process on ImageNet (divided by batch size).
  Abbreviations: ``Conv'': convolution; ``BN'': batch nomalization; ``FC'': fully connected layer;  ``EW-Add'': element-wise addition; ``F'': forward; ``B'': backward.}
    \label{tab:ops}%
    \resizebox{\columnwidth}{!}{
    \begin{tabular}{cccccc}
        \toprule
        Op Name         & Op Type  & ResNet18  & GoogleNet \\
        \midrule
        Conv (F)        & Mul\&Add & 1.88E+09  & 1.58E+09 \\
        Conv (B)        & Mul\&Add & 4.22E+09  & 3.05E+09 \\
        BN              & Mul\&Add & 3.06E+06  & 3.23E+06 \\
        FC              & Mul\&Add & 5.12E+05  & 1.02E+06 \\
        EW-Add (F)      & Add      & 7.53E+05  & 0 \\
        EW-Add (B)      & Add      & 9.28E+05  & 0 \\
        SGD Update (B)  & Mul\&Add & 1.15E+07  & 5.97E+06 \\
        \bottomrule
    \end{tabular}
    }
\end{table}

Besides the studies on improving inference efficiency, 
there are also some studies that focus on the training process. WAGE and FullINT~\citep{wage,fullint} implement fully fixed-point training with 8-bit and 16-bit integers to reduce the training cost. But they fail to achieve an acceptable accuracy since the dynamic range of data in training is large, and SGD algorithm needs small quantization error to ensure convergence. \textbf{This contradiction between the large dynamic range requirement of training algorithm and the small representation range of high-efficiency integer data format is the first challenge of low-bit training.} 

Floating point format has a larger representation range than fixed-point format with the same bitwidth.
%when the bit width is the same, 
FP8, HFP8 and S2FP8~\cite{fp8,hfp8,s2fp8} adopt 8-bit floating-point multiplications in convolution, in which more element-wise exponent bits are used to get a larger dynamic range. However, the precision of effective number is lost and they have to use complex quantization format. On the other hand, the dynamic range of intermediate accumulation results is too large and can only be conducted in the floating-point format, which results in higher energy consumption than integer accumulation. \textbf{How to realize low cost multiplications and accumulations (MACs) for high dynamic range floating-point data format is the second challenge of low-bit training.}
%But the energy efficiency of the convolution operation is the most crutial part in training process since it accounts for
%the majority of the operations, as shown in Table~\ref{tab:ops}. %and the convolution consists of multiplication and addition. 

\begin{figure}[tb]
\begin{center}
\vskip 0.2in
\includegraphics[width=0.9\columnwidth]{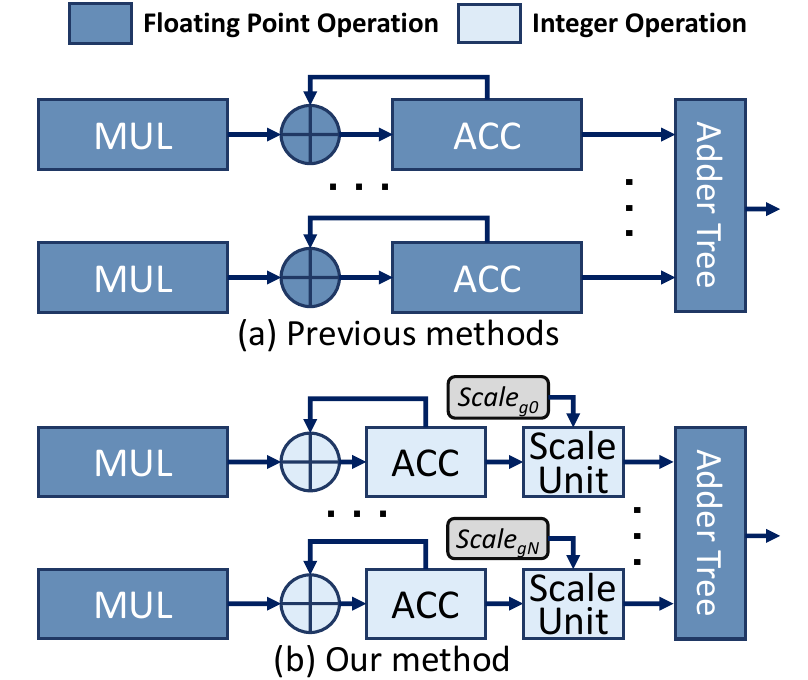}
\caption{The adder tree convolution hardware architecture. (a) Previous studies~\citep{hfp8} use low-bit floating-point multiplication (FP MUL) (e.g., 8-bit), but single precision accumulations are still needed. (b) We not only makes MUL less than 8-bit, but also simplifies the local accumulator.}% with a small overhead of scaling operation thanks to the special format of group-wise scaling factor.}
\label{pic:hardware-arch}
\vskip -0.2in
\end{center}
\end{figure}

In this work, we design a novel low-bit tensor format with multi-level scaling (MLS format) to maintain a high representation capability, which in the meantime could be manipulated by our customized hardware design with high energy efficiency. Specifically, in the MLS format: \textbf{1)} Element-wise scaling is used to improve the dynamic range of data representation, which greatly reduces the quantization error; \textbf{2)} A specially designed group scaling factor is used to reduce the element-wise exponent bits with smaller overhead, so that the accumulation can be simplified to integer accumulation without hurting the representational capability. Also, the specially designed group scaling could be conducted efficiently by shifting and additions instead of multiplications. Through these two techniques, we can reduce the dynamic range of most computing units while keeping the overall quantization error small, so as to achieve accurate and efficient calculation.

\begin{figure}[tb]
\vskip 0.2in
\begin{center}
\centerline{\includegraphics[width=0.96\columnwidth]{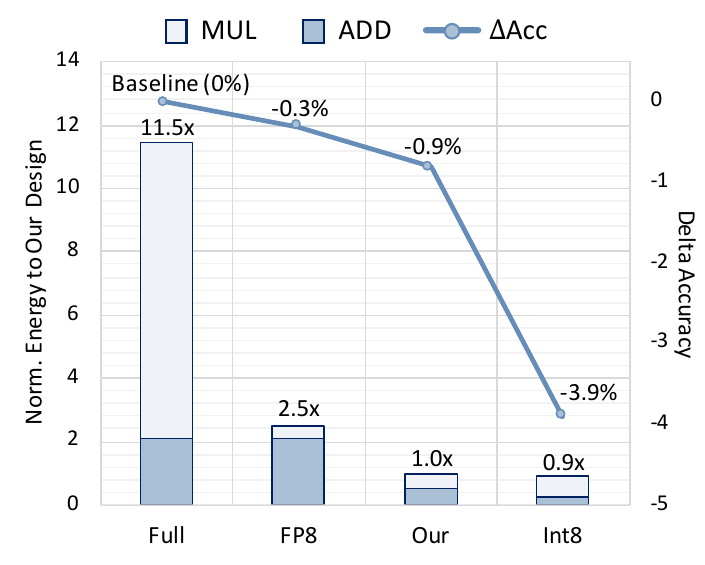}}
\caption{The model accuracy drop (ResNet-18 on ImageNet) and energy consumption of calculating 3$\times$3 convolutions with different training framework, nomalized to our design. FP8: \cite{hfp8}; Int8: \cite{fullint}.}
\label{fig:energy-illustration}
\end{center}
\vskip -0.2in
\end{figure}

Common hardware (e.g., GPU) is designed to support general floating-point arithmetic and can not support most of the existing low-bit tensor format efficiently. What's more, systolic array architecture that is widely used in NN accelerator treats convolution as general matrix multiplication and can not support data format with group-wise scaling. \textbf{Hence, the third challenge of low-bit training is that common hardware does not support specially designed data formats with group-wise scaling.} To tackle this challenge, we design \textbf{3)} the low-bit tensor convolution arithmetic unit with the MLS format to support our training framework efficiently. Our computing unit consists of low-bit multiplication, integer intra-group accumulation, group-wise scale unit, and inter-group addition tree, as shown in Fig.~\ref{pic:hardware-arch} (b). Different from previous methods with similar architecture (Fig.~\ref{pic:hardware-arch} (a)) and systolic array designs, our multiplications have smaller bit-width and the accumulations are conducted with fixed-point arithmetic instead of floating-point arithmetic. Therefore, as shown in Fig.~\ref{fig:energy-illustration}, our framework can largely reduce the energy consumption of MACs in convolution operations, compared with the full-precision and 8-bit floating-point frameworks.

To summarize, the contributions of this paper are:
%Utilizing the MLS format, one can strike a superior trade-off between accuracy and energy efficiency.
%Specifically, the multi-level scaling technique in the MLS format extracts the common exponent of tensor elements as much as possible to reduce the element-wise bit-width, while retaining the representational capability in the meantime. 

%On the other hand, all the multiplications and the majority of additions in the convolution operation could be conducted with low-bit floating-point or fixed-point arithmetic. %, thus boosting the energy efficiency of the training process.

%An illustrating comparison on the architecture and the energy consumption between our framework and existing ones is shown in Fig.~\ref{pic:hardware-arch} and Fig.~\ref{fig:energy-illustration}. To summarize, the contributions of this paper are:

 %Existing integer training frameworks simplify all the multiplications and additions but that hurts the model accuracy, and the low-bit floating-point training methods only simplify the multiplications but remain the others. 

%The total energy of CNN training can be shown in 
%Therefore, this work aims at simplifying convolution to low-bit operations, while retaining a similar performance with the full-precision baseline. The contributions of this paper are:

\begin{enumerate}
\item This paper proposes the \textbf{MLS tensor format} to strike a superior balance between representation capability and energy efficiency. %for the convolution operation. And we describe our 
\textbf{The element-wise scaling} improves the dynamic range of data representation, and by using \textbf{the group-wise scaling}, the element-wise exponent bitwidth can be kept low, so that the intra-group accumulation can be conducted with integer accumulation units for higher energy efficiency. We elaborate the corresponding low-bit training framework to leverage the MLS tensor format, and analyze that our MLS tensor format can be manipulated efficiently with our \textbf{low-bit tensor convolution arithmetic}. 

\item Experimental results demonstrate the representational capability of the MLS format: For training ResNets, VGG-16, and GoogleNet on CIFAR-10, using 1-bit mantissa and 2-bit exponent for each element can achieve an accuracy loss within $1\%$. For training these models on ImageNet, 4-bit mantissa and 2-bit exponent are adequate to achieve an accuracy loss within $1\%$.

\item We conduct \textbf{hardware design of low-bit tensor convolution arithmetic with MLS format}, and shows that our framework can indeed conduct MACs in convolutions efficiently, without degrading the model accuracy. We implement Register Transfer Language (RTL) designs of computing units with different arithmetic. And the energy consumption simulation shows that training a variety of models with our framework could achieve $8.3\sim 10.2\times$ and $1.9\sim 2.3\times$ higher energy efficiency than training with 32-bit and 8-bit~\cite{hfp8} floating-point arithmetic. On the other hand, we can achieve much higher accuracy than previous fixed-point training frameworks~\cite{wage,fullint} with comparable energy efficiency.
\end{enumerate}

\begin{figure}[tb]
\vskip 0.2in
\begin{center}
\centerline{\includegraphics[width=0.9\columnwidth]{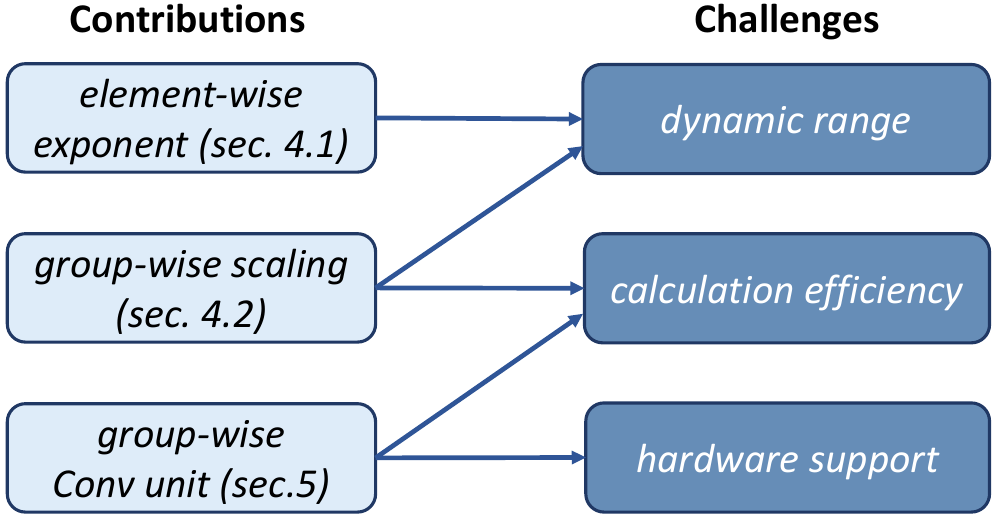}}
\caption{Three contributions of this paper and three challenges of low-bit training.}
\label{pic:contributions}
\end{center}
\vskip -0.2in
\end{figure}

The correspondences of the challenges in low-bit training and our contributions are summarized in Fig.~\ref{pic:contributions}. And the rest of this paper is organized as follows. Sec.~\ref{sec:related} discusses the related work of low-bit training, and Sec.~\ref{sec:Preliminary} gives basic knowledges on the training framework of CNNs. In Sec.~\ref{sec:dataformat}, we explain the reason why we proposed element-wise scaling and group-wise scaling in CNN low-bit training, and summarize them in MLS tensor format. The corresponding convolution arithmetic unit design is proposed in Sec.~\ref{sec:framework}. The dynamic quantization method and its overhead are discussed in Sec.~\ref{sec:dynamic_quant}, and the experiment results are shown in Sec.~\ref{sec:exp}. Finally, we draw the conclusion in Sec.~\ref{sec:conclusion}.

\section{Related work}
\label{sec:related}

%The studies on neural network quantization can be divided into three categories: 1) \textit{Post-training quantization} quantizes a pre-trained full-precision model without finetuning. 2) \textit{Quantization-aware training} considers the quantization effects into the training process to acquire a quantized model to accelerate the inference process. 3) \textit{Low-bit training} uses low-bit arithmetic in the training process. Our work belongs to the third category and aims at improving the energy efficiency of the low-bit training process.

\subsection{Post-Training Quantization}
\label{sec:post-train}
Earlier quantization methods like \citep{deepcompression} focus on the post-training quantization, and quantize the pre-trained full-precision model using the codebook generated by clustering or other criteria (e.g., SQNR~\cite{sqnr}, entropy~\cite{entropy}).
POST~\citep{post} and HAWQ~\citep{hawq} select the quantization bit-width and clipping value for each channel through the analytical investigation, but the per-channel precision allocation was not hardware-friendly. 
GEMMLOWP~\citet{gemmlowp} propose an integer arithmetic convolution for efficient inference, but it's hard to be used in training because the scale and bias of the quantized output tensor should be known before calculation.
MSFP\citet{msfp} proposes a new class of data formats developed for production cloud-scale inference on custom hardware.
These methods show that low-bit CNN models still have adequate representation ability. However, these methods are aming to accelerate inference of a pre-trained model and can not accelerate the training process. 
%These post-training quantization methods provide enlightening theoretical analysis on quantization error, and show that CNN models with reduced bit-width still have adequate representation ability. 
%However, their quantization process are based on pretrained models, and cannot accelerate the training process. 

\subsection{Quantization-Aware Training}
Quantization-aware training considers quantization effects in the training process to further improve the accuracy of the quantized model. %Some methods train an ultra low-bit network like 
It is used for training binary~\citep{xnor} or ternary~\citep{twn} networks. Despite that the follow-up studies~\citep{reactnet}\citep{irnet} have proposed new techniques to improve the accuracy, the extremely low bit-width still causes notable accuracy degradation. 
Other methods seek to retain the accuracy with relatively higher precision, e.g., 8-bit~\citep{quantizetraining}.
\citet{ristretto} develop GPU-based training framework to get dynamic fixed-point models or Minifloat models. \citet{pact} parameterizes and trains the clipping value in the activation function to properly restrict the range of activation. 
These methods obtain quantized models that achieve better trade-off of accuracy and inference efficiency, but the training process is still full-precision. %and running on GPUs. %using floating-point arithmetic.

\subsection{Low-Bit Training}
\label{sec:low bit train}
%Some studies~\citep{flexpoint} %like~\cite{Shifted, floatfix} 
%seek to use flexible data format for better representation ability. 
%Such as 
To accelerate the training process, \citet{dorefa} propose to use fixed-point arithmetic in both the forward and backward processes.
\citet{wage, fullint} implement full-integer training frameworks for integer-arithmetic machines. However, these methods cause notable accuracy degradation. \citet{rangebn} use 8-bit and 16-bit integer arithmetic~\citep{gemmlowp} and achieve a better accuracy. But this arithmetic~\citep{gemmlowp} is designed for accelerating inference and requires knowing the output scale before calculation. Therefore, although \citet{rangebn} quantize the gradients in the backward process, it is not practical for actual training acceleration. 
To summarize, full-integer training frameworks have high energy efficiency, but still suffer from large accuracy degradation when the bit-width is reduced to 8 bit.
%Integer training methods shows high energy efficiency but the accuracy drop is often large when the bit-width is reduced to 8-bit.

Besides the studies on full-integer training frameworks, some studies propose new low-bit formats. BFloat~\citet{bf16} use a 16-bit floating-point format that is more suitable for CNN training. Flexpoint~\citet{flexpoint} propose the format that contains 16-bit mantissa and 5-bit tensor-shared exponent (scale), which is similar to the dynamic fixed-point format~\citet{dfp16}.
Recently, 8-bit floating-point formats~\citep{fp8,hfp8,s2fp8} are used with chunk-based accumulation. However, to ensure a sufficient representation range, 
the exponent bit-width in their format is larger than 5, which makes the operations (especially the accumulation) using these formats inefficient. More recently, a radix-4 data format \citep{ultra} is proposed along with two-stage quantization to realize 4-bit training, %utilize 4-bit training, 
but the accuracy is not satisfying enough and its computation is complex. 
In this work, the MLS tensor format is designed to have a small exponent bit-width, such that the accumulation can be conducted using fixed-point arithmetic, while retaining the overall model accuracy.
%remaining the overall model accuracy. %to strike a better trade-off between accuracy and energy efficiency.
%however, the trade-off of the accuracy and bit-width (efficiency) is still 
%some stills suffers from notable accuracy degradation or still resort to less efficient high bit-width / floating-point data. 
%explore the keys to Lower-bitwidth quantized training. 

\section{Preliminary}
\label{sec:Preliminary}

\subsection{Computation Flow for CNN Training}
\begin{figure}[tb]
\begin{center}
%\framebox[4.0in]{}
%\fbox{\rule[-.5cm]{0cm}{4cm} \rule[-.5cm]{4cm}{0cm}}
\centerline{\includegraphics[width=\columnwidth]{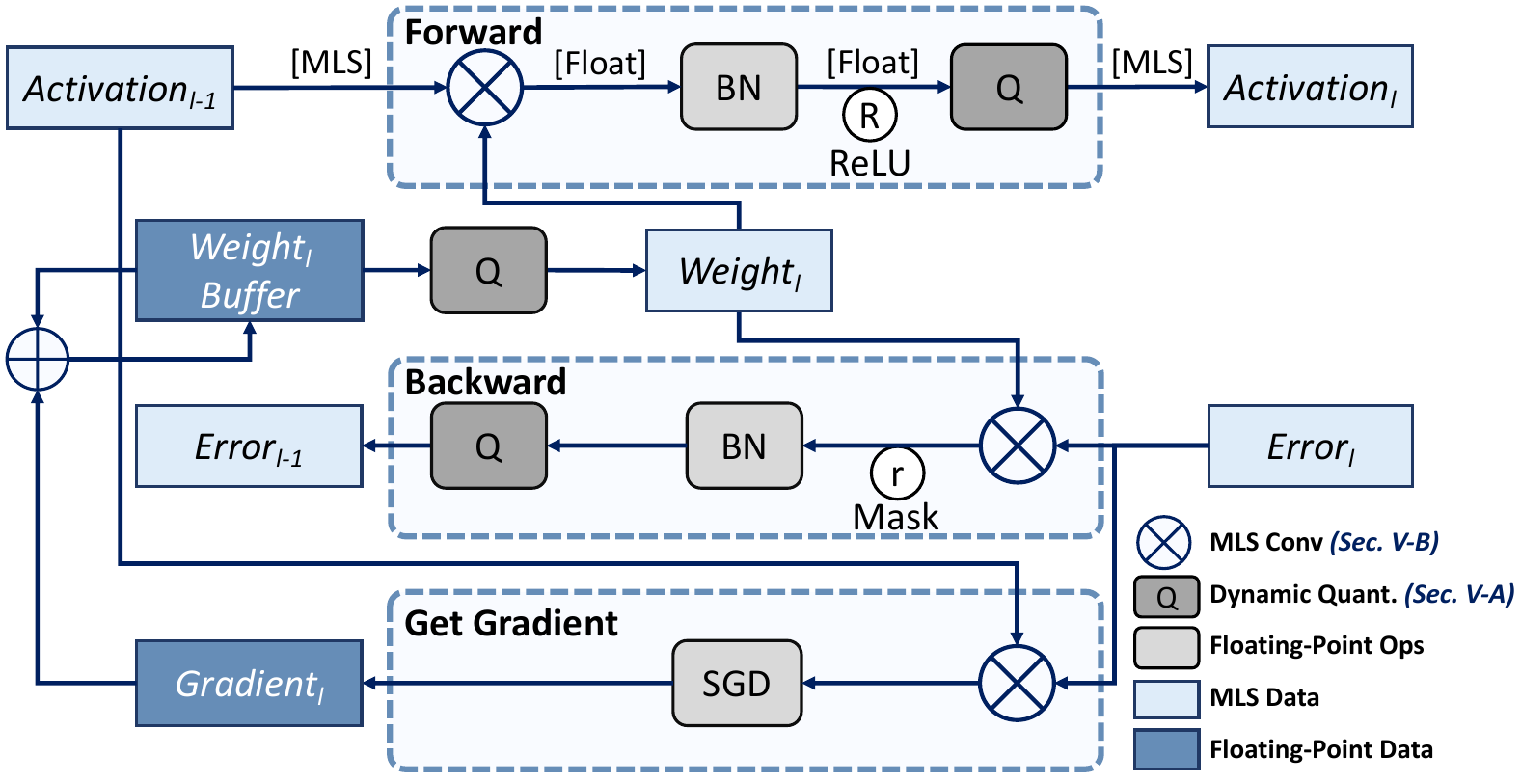}}
\caption{Computation flow of our low-bit training framework.} %\todo{change the section number in figure}} %\zksays{Need to change larger or one column, and change the legend to MLS Data(Sec.3),Float Data,Float Ops  and emphasize it's not quantization-aware training.}}%Floating-point data are quantized to low-bit before the convolution operations.}
\label{train-framework}
\end{center}
\end{figure}

In this work, we denote the filter coefficient and feature map of convolution as weight and activation, respectively. In the back-propagation, the gradient of convolution results and weights are denoted as error and gradient, respectively. As shown in Fig.~\ref{train-framework}, generally, in a convolutional layer, convolution is followed by batch normalization (BN), nonlinear activation (ReLU is used in this work) and other operations like pooling.

As shown in Table~\ref{tab:ops}, the MACs in convolutions are the majority of the operations in a convolution layer. Hence, conducting the MACs with low-bit arithmetic in convolutions can boost the energy efficiency of the training process.
And conducting other operations (e.g., BN, weight update) using high bit-width helps to stablize training and make the accuracy higher. %and remaining other operations like batch normalization and weight update in higher bit-width makes it easy to keep the training stable and lead to higher accuracy. 
Therefore, we focus on the quantization before all three types of Conv (Conv of weight and activation, weight and error, activation and error). And the output data of Conv is in floating-point format for other operations like BN.

\subsection{Basic Formula of Convolution}
Weight, activation, and error are all 4-dimension tensors in the training process. For activation and error, the four dimensions are sample in batch ($N$), channel ($C$), feature map height ($F_h$), and feature map width ($F_w$). For weight, the four dimensions are output channel ($C_o$), input channel ($C_i$), kernel height ($K_h$), kernel width ($K_w$).

We take Conv(Weight, Activation) ($Conv(\bm{W}, \bm{A})$) as the example to introduce the basic formula of convolution between two 4-dimension tensors in training, and the other two types of convolution can be implemented similarly.
Denoting the input channel number as $C$ and the kernel size as $K=K_h=K_w$, the original formula of convolution is:

\begin{equation}
\begin{split}
  % \etZ_{n,co,x,y} = \sum_{ci=0}^{Ci-1} \sum_{i=0}^{K-1} \sum_{j=0}^{K-1} \etW_{co,ci,i,j} \times \etA_{n,ci,x+i,y+j}
      \bm{Z}[n,co,x,y] &= Conv(\bm{W}, \bm{A}) = \sum_{ci=0}^{C-1} \sum_{i=0}^{K-1} \sum_{j=0}^{K-1} \\
      & \bm{W}[co,ci,i,j] \times \bm{A}[n,ci,x+i,y+j]
\end{split}
\label{equ:conv}
\end{equation}

We can see that every element in the output 4-dimension tensor is calculated by three loops of MACs. And three dimensions of input tensors are included in this accumulation. In common training frameworks based on hardware platforms like TPU and GPU, these tensors are processed with the ``image to column'' transformation. Then the convolution is calculated as a general matrix multiplication, in which grouping techniques cannot be used~\citep{tpu,tensorflow,pytorch}. But in many customized CNN accelerators~\citep{deeper,angle}, parallel PE units and addition tree architecture are used. %\todo{The advantage of this architecture is that the each loop can find the corresponding hardware.?} 
The MACs can be grouped into intra-group ones and inter-group ones,
%to divide the accumulations into intra-group and inter group,
which makes it possible for us to apply group-wise scaling. Next, we will show the advantages of group-wise scaling through data format design and hardware design.

\section{Mulit-level Scaling Low-bit Tensor Format}
\label{sec:dataformat}

%\todo{Let's introduce intra/inter-group accumulation before this! And analyze that the element-wise bitwidth is the ``operating'' bitwidth of hardware units, and is crucial to energy efficiency.}

Using low-bit arithmetic in the training process is beneficial for the energy efficiency. However, retaining a good accuracy in a low-bit fixed-point training process is challenging, since that the backpropagated gradients need high precision ~\cite{dorefa}. In this work, we design a MLS low-bit tensor format to retain the representational power of low-bit representations in CNNs. It consists of three levels of scaling factors: 1) Tensor-wise scaling factor; 2) Group-wise scaling factor; 3) Element-wise exponent. By incorporating the multi-level scaling technique, the element-wise bitwidth can be largely reduced to boost the energy efficiency, while the overall dynamic range is preserved. 

In this section, we give the design details of the MLS low-bit tensor format, which is the core of our low-bit training framework. And in the next section Sec.~\ref{sec:framework}, we will elaborate on the framework and hardware design centering around the MLS format, to demonstrate that the conversion and computation of the MLS-formatted data are energy efficient.
%In this section, 

%We denote the filter and feature map of convolutions as weight and activation, and the gradients of feature map and weights are denoted as error and gradient, respectively. And all these tensors are 4-dimension tensors in the training process.%W, A, and E of each layer are 4-dimension tensors in training. 
%For A and E, the four dimensions are sample in batch (N), channel (C), feature map height (H), feature map width (W). And output channel (Co), input channel (Ci), kernel height (K), kernel width (K) for W.

\subsection{Overall Mapping Formula of the MLS Format}
\label{sec:mls_mapping}
%\todo{need explanation of i,j,k,l; move some content of 3.3 here!}

\begin{figure}[ht]
\vskip 0.2in
\begin{center} 
\centerline{\includegraphics[width=\columnwidth]{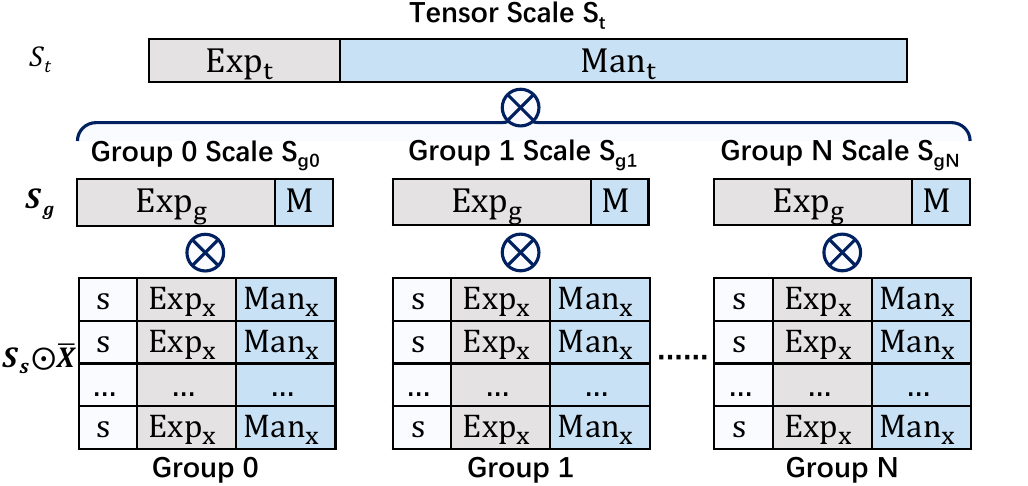}}
\caption{The multi-level scaling (MLS) low-bit tensor format.} 
\label{pic:data-format}
\end{center}
\vskip -0.2in
\end{figure}

\begin{figure*}[tb]
\vskip 0.2in
\begin{center}
\centerline{\includegraphics[width=0.9\linewidth]{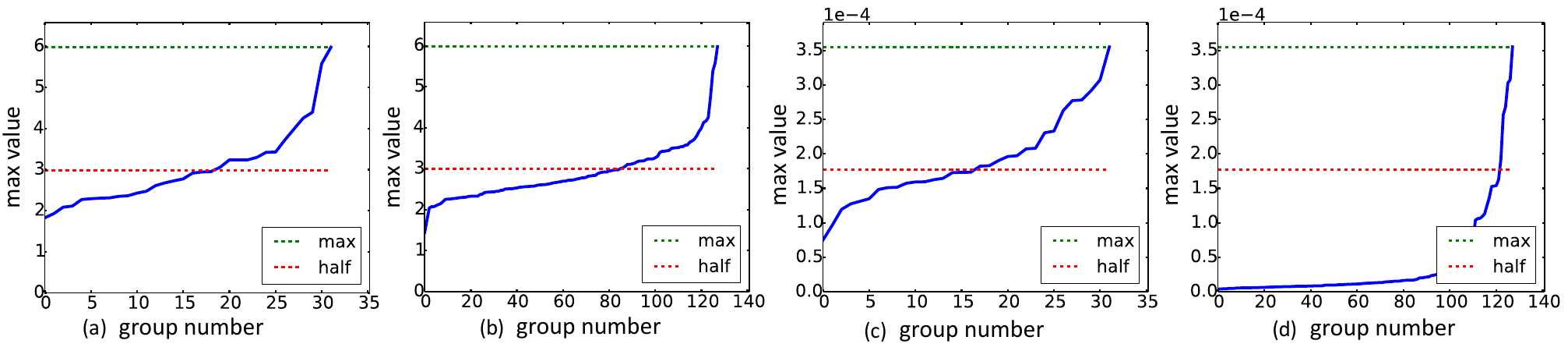}}
\caption{Maximum value of each group of activation (a, b) and error (c, d). (a)(c): Grouped by channel; (b)(d): Grouped by sample.}
\label{pic:max-by-group}
\end{center}
\vskip -0.2in
\end{figure*}

%\todo{sec title Mapping Formula of the MLS Tensor Format}fin our
%In quantized CNNs, a floating-point value is quantized to a quantized value, which can be represented using some fixed-point values.

%In quantized CNNs, floating-point values are quantized to use the fixed-point representation.

%a combination of fixed-point values with the help of auxiliary parameters.
% The common relationship between the floating-point number and the quantized element is:
In a commonly used scheme~\citep{gemmlowp}, the mapping %of the fixed-point numbers to the represented floating-point number is
function from fixed-point representation and the floating-point values is
$ float = scale \times (Fix + Bias) $, in which $scale$ and $Bias$ are shared in one tensor. %and chosen to minimize the quantization error.
In training, however, 
since data distribution changes over time, one cannot simplify the $Bias$ calculation as they do.
Thus, we adopt an unbiased quantization scheme, % in our low-bit training framework,
% Since tensors in training process are 4-demontional, we extend the scaling factor to three levels, and propose \textbf{Multi-Level Scaling low-bit tensor format (MLS format)}.
and extend the scaling factor to three levels for better representation ability. The resulting MLS tensor format is illustrated in Fig.~\ref{pic:data-format}.
Denoting a 4-dimensional tensor that is the operand of Conv (weight, activation, or error) as $\bm{X}$, %the mapping formula of the quantization scheme to the MLS tensor format is
the mapping formula of the MLS tensor format is
%Its components are shown in Fig.~\ref{pic:data-format}, and the basic formula is: %The first two are tensor-wise scaling and group-wise scaling, the basic formula of this format is:
\begin{equation}
  % {\etA_f}_{i,j,k,l} = {\etS_s}_{i,j,k,l} \times S_t \times {\emS_g}_{i,j} \times {\etA}_{i,j,k,l}
      \bm{X}[i,j,k,l] = {\bm{S_s}}[i,j,k,l] \times S_t \times {\bm{S_g}}[i,j] \times \bar{\bm{X}}[i,j,k,l]
    \label{equ:basic}
  \end{equation}
where $[\cdot]$ denotes the indexing operation, $\bm{S_s}$ is a 1-bit sign tensor (``s'' in Fig.~\ref{pic:data-format}), $S_t$ is a full-precision tensor-wise scaling factor, and $\bm{S_g}$ are group-wise scaling factors shared in each group. $\bm{S_g}$ and $\bar{\bm{X}}$ use the same data format, which we refer to as $\langle E, M\rangle$, a customized floating-point format with E-bit exponent and M-bit mantissa (no sign bit). %A  value of a number in format
A value in the format $\langle E, M\rangle$ is %can be represented as: 
\begin{equation}
\begin{split}
    float &= I2F(Man, Exp) = Frac \times 2^{-Exp} \\
          &=\left(1+\frac{Man}{2^M}\right)\times 2^{-Exp}
\end{split}
    \label{equ:em-float}
\end{equation} 
where $Man$ and $Exp$ are the M-bit mantissa and E-bit exponent, and $Frac \in [1, 2)$ is a fraction.

%where $\bm{S_s}$ is the 1-bit sign tensor, $S_t$ is tensor-wise scaling shared across elements in the whole tensor, and $\bm{S_g}$ is only shared across elements in one group which is a subset of the tensor. There are different ways of grouping, we consider three natural grouping methods: grouping by the first dimension of tensor, the second dimension of tensor, or the first and the second dimensions at the same time.

\subsection{Group-wise Scaling}
%\todo{Add this to the discussion of group-wise scaling}

The dynamic ranges of weight, activation and error are large in training, but we find that these values are not evenly distributed. The values in different groups have distinct dynamic ranges, as shown in Fig.~\ref{pic:max-by-group}. The blue line shows the max value in each group when activation and error are grouped by channel or sample. If we use the overall maximum value (green lines in Fig.~\ref{pic:max-by-group}) as the overall scaling, many small elements will be swamped. And usually, there are over half of the groups, in which all elements are smaller than half of the overall maximum (red line). Thus, to fully exploit the bit-width, it is natural to use group-wise scaling factors. Our work considers three grouping dimensions: 1) the 1-st dimension of tensor, 2) the 2-nd dimension of tensor, or 3) the 1-st and the 2-nd dimensions simultaneously.
%\todo{Since we emphasize that one of the distinct challenge of low-bit training compared with low-bit inference is the dynamic range of error, we should plot the error value distribution too}

Naive floating-point group-wise scaling in previous studies~\cite{xnor} cannot bring actual hardware acceleration. Since when the values of different groups are accumulated, the floating-point scaling factors need to be multiplied back to low-bit elements, which involves floatint-point multiplications. 

To facilitate a hardware-friendly low-bit training framework, we propose a special scaling format, the floating-point group-wise scaling is separate into tensor-wise and group-wise scaling factors. The first level \textbf{tensor-wise scaling factor} $S_t$ is an ordinary floating-point number ($\langle E_t,M_t\rangle = \langle 8, 23\rangle$), to retain the precision as much as possible. 
%such that the group-wise scaling could be implemented efficiently with shifting and additions.

Considering the actual hardware implementation cost, there are some restrictions on the second level \textbf{group-wise scaling factor} $\bm{S_g}$.
%cannot be set to ordinary floating-point format.
%\todo{overhead not come from equation 2 I2F, but come from group-wise scaling multiplication another number, equation is not a real I2F but a data format which will not be calculated in practice, so this part may be rewritten}.
Since calculation results of different groups need to be aggregated, using $\bm{S_g}$ in an ordinary floating-point format leads to expensive conversions in the hardware implementation. %and operations necessary 
%otherwise it will introduce additional overhead in convolution, which makes the low-bit training not get much benefit. 
Hence, we propose two special hardware-friendly group-wise scaling schemes, whose formats can be denoted as $\langle E_g, 0\rangle$, and $\langle E_g, 1\rangle$, respectively.  
The scaling factor in $\langle E_g,0\rangle$ format is simply a power of two, which can be implemented easily as shifting on the hardware. 
%Bring $M_g=1$ into Eq.~\ref{equ:em-float}, $\langle E_g,1\rangle$ scaling factor can be represent as:
From Eq.~\ref{equ:em-float}, a $S_g=I2F(Man_g, Exp_g)$ value in the $\langle E_g, 1\rangle$ format can be written as
\begin{equation}
\begin{split}
    S_g &= \left(1+\frac{Man_g}{2}\right)\times 2^{-Exp_g} \\
    &= 
    \begin{cases}   
        2^{-Exp_g} + 2^{-Exp_g-1}   & Man_g=1 \\     
        2^{-Exp_g}                  & Man_g=0     
    \end{cases}
\end{split}
    \label{equ:group-scaling}
\end{equation} 
which is a sum of two shifting, and can also be implemented with low hardware overhead. We will see that MLS tensor convolution arithmetic benefits from group-wise scaling factor's special format with very few mantissa bits in Sec.~\ref{sec:mls-conv} (Eq.~\ref{eq:groupwise_scale}). 

\subsection{Element-wise Scaling}
The third level scaling factor $S_x=I2F(0, Exp_x) =2^{-Exp_x}$ is the \textbf{element-wise exponent} in $\bar{X}=S_x (1+\frac{Man_x}{2})$, %which is a power of two: $S_e = $, 
and we can see that the elements of $\bar{\bm{X}}$ in Eq.~\ref{equ:basic} are in a $\langle E_x, M_x\rangle$ format. The specific values of $E_x$ and $M_x$ determine the cost of the MAC operation, which will be discussed in Sec.~\ref{sec:mls-conv}. Compared with integer data format ($E_x=0$), adding element-wise exponent helps achieve a balance in the dynamic range and precision of representation. %The reason why we list it out is that when used with other scaling, the exponent will have a different offset from the general floating-point number, and this analysis method allows the bit-width of the element-wise exponent to be zero, which represents a fixed-point format.
And by using group-wise scaling, the bit-width of $\bm{\bar{X}}$ can be largely reduced. %What differentiates our work from previous studies that utilize scaling factors is that the formats of the multi-level scaling factors are carefully designed to avoid all floating-point operations in convolutions. Unlike previous studies~\cite{xnor, rangebn}, in our work, all the scaling operations, the intra- and inter-group MACs can be implemented efficiently. 
%Moreover, the inter-group accumulation needs to be conducted using floating-point additions. , and the inter-group accumulation could be conducted with integer arithmetic.

\section{Low-bit Training Framework}
\label{sec:framework}
In this section, we describe the low-bit training framework to leverage the MLS tensor format. A training iteration in our low-bit training framework is summarized in Alg.~\ref{alg:low-bit-training}, and the computation flow of one layer is shown in Fig.~\ref{train-framework}. Note that, our framework is different from a quantization-aware training framework in that the convolution operands are actually quantized to the low-bit MLS format in our computation flow. In the backward propagation (Alg.~\ref{alg:low-bit-training}, line 13), we use the update formula of the vanilla stochastic gradient descent (SGD) for clarity, whereas in practice, one can use other optimizers such as SGD with momentum. The $t$ subscripts denoting the time step $t$ are all omitted for simplicity. 

In this section, we will describe two core parts of the framework to demonstrate why the conversion and computation of the format are energy efficient: Sec.~\ref{sec:dynamic_quant} describes the dynamic quantizaiton $DynamicQuantization$, and Sec:~\ref{sec:mls-conv} describes the low-bit tensor convolution arithmetic $LowbitConv$. They are actually FP-to-MLS conversion and the MLS MLS-to-FP conversion in the training framework.

\begin{algorithm}[tb]
\caption{The low-bit training framework}
\label{alg:low-bit-training}
% \hspace*{0.02in} {\bf Input:}
% number of layers: $L$, float weights: $\tW^{1:L}_{t}$, float input: $\tA^0$, label: $\mT$, learning rate: $r$\\
% \hspace*{0.02in} {\bf Output:}
% new float weights: $f\tW^{1:L}_{t+1}$ \\
% \hspace*{0.2in} \textbf{forward propagation}
\LinesNumbered
\DontPrintSemicolon
\SetNoFillComment 
\KwIn{
  $L$: number of layers;
  $\bm{W^{1:L}}$: current float weights;
  $\bm{A^0}$: inputs;
  $\bm{T}$: label;
  % label: $\bT$,
  $lr$: learning rate}
\KwOut{
$\bm{W^{1:L}_{t+1}}$: updated float weights for the next step $t+1$}
%\begin{algorithmic}[l]
%  \NoNumber{\textbf{// forward propagation}}
\nonl $\empty$ \\
\nonl \textbf{/* forward propagation */}~\\
\For{$l$ in $1:L$}{%\\
    $q\bm{W^{l}} = DynamicQuantization(\bm{W^{l}})$~\\
$q\bm{A^{l-1}} = DynamicQuantization(\bm{A^{l-1}})$~~\\
$\bm{Z^{l}} = LowbitConv(q\bm{W^{l}},q\bm{A^{l-1}})$~\\
$\bm{Y^{l}} = BatchNorm(\bm{Z^{l}})$~\\
$\bm{A^{l}} = Activation(\bm{Y^{l}})$
}

$\frac{\partial{loss}}{\partial{\bm{A^{l}}}} = Criterion(\bm{A^{L}},\bm{T})$~\\
%\end{algorithmic}
%\hspace*{0.2in}  \textbf{backward propagation}  
% \begin{algorithmic}[1]
\nonl $\empty$ \\
\nonl {\textbf{/* backward propagation */}}~\\
\For{$l$ in $L:1$}{
     $\frac{\partial{loss}}{\partial{\bm{Y^{l}}}} = \frac{\partial{loss}}{\partial{\bm{A^{l}}}} \times Activation'(\bm{Y^{l}})$~\\
     $\frac{\partial{loss}}{\partial{\bm{Z^{l}}}} = \frac{\partial{loss}}{\partial{\bm{Y^{l}}}} \times \frac{\partial{\bm{Y^{l}}}}{\partial{\bm{Z^{l}}}}$~\\
     $q\bm{E^{l}} = DynamicQuantization(\frac{\partial{loss}}{\partial{\bm{Z^{l}}}})$~\\
     $\bm{G^{l}} = LowbitConv(q\bm{E^{l}}, q\bm{A^{l-1}})$~\\
     $\bm{W^{l}_{t+1}} = \bm{W^{l}} - lr \times \bm{G^{l}}$~\\
    \If{$l$ is not $1$}{
       $\frac{\partial{loss}}{\partial{q\bm{A^{l-1}}}} = LowbitConv(q\bm{E^{l}}, q\bm{W^{l}})$~\\
       $\frac{\partial{loss}}{\partial{\bm{A^{l-1}}}} = STE(\frac{\partial{loss}}{\partial{q\bm{A^{l-1}}}})$
}
}
\nonl {\bfseries Return} $\bm{W^{1:L}_{t+1}}$
\end{algorithm}

\begin{algorithm}[tb]
    \caption{Dynamic Quantization}
%\algorithmfootnote{$\dagger$: 
%The underflow handling follows the IEEE 754 standard~\citep{hough2019ieee}.}
%\footnote{The underflow handling follows the IEEE 754 standard~\citep{hough2019ieee}.}
    \label{alg:dynamic-quantization}

% \LinesNumbered
%   \everypar={\nl}

%  \SetAlgoLined
\DontPrintSemicolon
\SetNoFillComment 
\KwIn{
    $\bm{X}$: float 4-d tensor;
    $\bm{R}$: $U[-\frac{1}{2}, \frac{1}{2}]$ distributed random tensor;
    $\langle E_g, M_g\rangle$: bit-width of group-wise scaling factors; $\langle E_x, M_x\rangle$: bit-width of each element
}
\KwOut{
    $\bm{S_s}$: sign tensor; $S_t$: tensor-wise scaling factor; $\bm{S_g}$: group-wise scaling factors; $\bm{\bar{X}}$: quantized elements
}
\nonl $\empty$ \\
\nonl \textbf{/* calculating scaling factors */} \\
$\bm{S_s} = Sign(\bm{X})$ \\
 $\bm{S_r} = GroupMax(Abs(\bm{X}))$ \\
 $S_t = Max(\bm{S_r})$  \\
 $\bm{S_{gf}} = \mS_r\div S_t$ \\
 $\bm{Exp_g}, \bm{Frac_g} = Exponent(\bm{S_{gf}}), Fraction(\bm{S_{gf}})$\\
 $\bm{Exp_g} = Clip(\bm{Exp_g}, 1-2^{E_g},0)$ \\
 $\bm{Frac_{g}} = Ceil(\bm{Frac_g}\times 2^{M_g}) \div 2^{M_g}$ \\
 $\bm{S_g} = \bm{Frac_{g}} \times 2^{\bm{Exp_g}}$ \\
\nonl $\empty$ \\
\nonl \textbf{/* calculating elements */}\\
%\end{algorithmic}
%\hspace*{0.2in} \textbf{rouding elements}
%\begin{algorithmic}[1]
 $\bm{X_f} = Abs(\bm{X}) \div \bm{S_g} \div S_t$ \\
 $\bm{Exp_x}, \bm{Frac_x} = Exponent(\bm{X_f}), Fraction(\bm{X_f})$\\

%\nonl \textbf{/* quantize $\bm{Frac_x}$ to $M_x$ bits with underflow handling$^\dagger$ */}
\nonl \textbf{// quantize $\bm{Frac_x}$ to $M_x$ bits with underflow handling}\\%$^\dagger$}\\%\footnotemark */\\
 $E_{xmin} = 1-2^{E_x}$\\
 $\bm{Frac_{xs}} = \bm{Frac_{x}}\times 2^{M_x}$ if not underflow, else  $\bm{Frac_x}\times2^{M_x-E_{xmin}+E_x}$ \\ %\tcp*{Scale the fraction}\\
 $\bm{Frac_{xint}} = Clip(SRound(\bm{Frac_{xs}}, \bm{R})),0, 2^{M_x}-1)$ \\%\tcp*{Quantize by stochastic rounding}\\
 $\bm{Frac_{x}} = \bm{Frac_{xint}}\times 2^{-M_x}$ if not underflow, else $\bm{Frac_{xint}}\times 2^{-M_x+E_{xmin}-E_x}$ \\ %\tcp*{Scale back to get the quantized fraction} \\
 $\bm{Exp_x} = Clip(\bm{Exp_x}, E_{xmin}, -1)$\\
 $\bar{\bm{X}} = \bm{Frac_{x}} \times 2^{\bm{Exp}_{x}}$ \\
\nonl {\bfseries Return} $\bm{S_s}$, $S_t$, $\bm{S_g}$, $\bar{\bm{X}}$
\end{algorithm}

\subsection{Dynamic Quantization to MLS Tensor}
\label{sec:dynamic_quant}

The dynamic quantization converts a floating-point tensor to a MLS tensor. There are two main steps, calculating the scaling factors $\bm{S_s}, S_t, \bm{S_g}$ and getting the quantized elements $\bar{\bm{X}}$, as shown in Alg.~\ref{alg:dynamic-quantization}. In Alg.~\ref{alg:dynamic-quantization}, the sign tensor, overall maximun and group-wise maximums are got firstly in line 1$\sim$3. And group-wise scaling factors are quantized by group-wise maximums in line 4$\sim$8. 
$Exponent(\cdot)$ and $Fraction(\cdot)$ are to obtain the Exponent (an integer) and Fraction (an integer represent numberts $\in[1,2)$) of a floating-point number, which are used in the quantization of group-wise scalings and element-wise numbers in line 5 and 10. The underflow handling follows the IEEE 754 standard~\citep{hough2019ieee} as shown in line 11$\sim$15.
When calculating the quantized elements $\bar{\bm{X}}$, we apply the stochastic rounding~\citep{stochastic} $SRound(x,r)$ as shown in line
13. It is implemented with a uniformly distributed random tensor $r \sim U[-0.5, 0.5]$ which can be generated offline as how it is done on GPU.
\begin{equation}
\begin{split}
    S&Round(x,r) = NearestRound(x+r) \\
                        &=
    \begin{cases}
    \lceil x \rceil     & \text{with probability $x - \lfloor x \rfloor$} \\
    \lfloor x \rfloor   & \text{with probability $\lceil x \rceil - x$}
    \end{cases}
\end{split}
\end{equation}%where ``w.p.'' denotes ``with probability''.%It is worth noting that

Note that Alg.~\ref{alg:dynamic-quantization} describes how we simulate the dynamic quantization process on floating-point platform. While in the hardware design, the exponent and mantissa are obtained directly, while the $Clip$ operations are conducted by taking out some bits from a machine number. %Thus, the actual cost mainly comes from the maximum value statistics $Max(\cdot)$ and the division in Alg.~\ref{alg:dynamic-quantization} Line 9. 

%can also be intercepted directly on the binary number. So the actual cost mainly comes from the statistic of maximum values and the division on line 1 of rounding elements.
%In the practice, the exponent and mantissa can be obtained directly, and the quantization operation can also be intercepted directly on the binary number. So the actual cost mainly comes from the statistic of maximum values and the division on line 1 of rounding elements.

\subsection{Low-bit Tensor Convolution Arithmetic}
\label{sec:mls-conv}
%\todo{For more clear notation, change tensor index subscripts to [i,j,k,l]}
In this section, we describe how to do convolution with two low-bit MLS tensors.
%Another important problem we need to deal with is how to do convolution with two multi-level scaling low-bit tensors. 
%(i.e., Conv(W, E), Conv(A,E)) 

%formula with only index difference.
%As we introduce in Sec.~\ref{sec:dataformat}, the relationship of the float tensor and the quantized low-bit tensor is as Eq.~\ref{equ:basic}. So 
Using the MLS tensor format and denoting the corresponding values (scaling factors $\bm{S}$, exponents $\bm{Exp}$ and fractions $\bm{Frac}$ in the following equations) of $\bm{W}$ and $\bm{A}$ by the superscript $^{(w)}$ and $^{(a)}$, one output element $\bm{Z}[n,co,x,y]$ of $Conv(\bm{W}, \bm{A})$ is calculated as:
\begin{equation} 
\begin{medsize}
\begin{split}
\bm{Z}[n,co,x,y] &= \sum_{ci=0}^{C-1} \sum_{i=0}^{K-1} \sum_{j=0}^{K-1}
            \left({S_{t}^{(w)}} {\bm{S_{g}^{(w)}}}[co,ci] \bar{\bm{W}}[co,ci,i,j]\right)\\
            &\qquad \qquad \left({S_{t}^{(a)}} \bm{S_{g}^{(a)}}[n,ci] \bar{\bm{A}}[n,ci,x+i,y+j]\right)\\
        &=  \left({S_{t}^{(w)}} {S_{t}^{(a)}}\right) \sum_{ci=0}^{C-1} 
           [ \left({\bm{S_{g}^{(w)}}}[co,ci] {\bm{S_{g}^{(a)}}}[n,ci]\right) \\
        &\quad  \sum_{i=0}^{K-1}\sum_{j=0}^{K-1}\bar{\bm{W}}[co,ci,i,j]\bar{\bm{A}}[n,ci,x+i,y+j] ] \\
        &= S_{t}^{(z)}\sum_{ci=0}^{C-1}{\bm{S^{(p)}}}[n,co,ci] \bm{P}[n,co,ci]
        \label{eq:Z_convwa}
% \etZ_{n,co,x,y} &= \sum_{ci=0}^{C-1} \sum_{i=0}^{K-1} \sum_{j=0}^{K-1} 
%             \left({S_{tw}}\times{\emS_{gw}}_{co,ci}\times\bar{\bm{W}}_{co,ci,i,j}\right)\times
%             \left({S_{ta}}\times{\emS_{ga}}_{n,ci}\times \bar{\bm{A}}_{n,ci,x+i,y+j}\right)\\
%         &=  \left({S_{tw}}\times {S_{ta}}\right) \times \sum_{ci=0}^{Ci-1} 
%             \left({\emS_{gw}}_{co,ci}\times{\emS_{ga}}_{n,ci}\right) \times
%             \sum_{i=0}^{K-1}\sum_{j=0}^{K-1}\bar{\bm{W}}_{co,ci,i,j}\times\bar{\bm{A}}_{n,ci,x+i,y+j}\\
%         &= {S_{tz}}\times\sum_{ci=0}^{Ci-1}{\bm{S}_{gp}}[n,co,ci]\times \etP_{n,co,ci,x,y}
\end{split}
\end{medsize}
\end{equation}
Eq.~\ref{eq:Z_convwa} shows that the accumulation consist of \textbf{intra-group} MACs that calculates $\bm{P}[n,co,ci]$ %(multiplication and accumulation) 
and \textbf{inter-group} MACs that calculates $\bm{Z}$. \\
 %The former one can be regarded as matrix multiplication, and we %expand the floating-point number according to 
%bring Eq.~\ref{equ:em-float} in to get the following integer calculation process:

\noindent\textbf{Intra-group MACs} The intra-group calculation of $\bm{P}[n,co,ci]$ is:

\begin{equation} 
\begin{medsize}
\begin{split}
% \bm{P}[n,co] = \sum_{i=0}^{K^2-1} \bm{W}[co,i]\times \bm{A}[n,i]
%             &= \sum_{i=0}^{K^2-1} \left({\emF_{w}}_{co,i}\times2^{{\emE_{w}}_{co,i}}\right)\times 
%             \left({\emF_{a}}_{n,i}\times 2^{{\emE_{a}}_{n,i}}\right)\\
%             &= \sum_{i=0}^{K^2-1}
%             \left({\emF_{w}}_{co,i}\times{\emF_{a}}_{n,i}\right)\times
%             2^{\left({\emE_{w}}_{co,i} + {\emE_{a}}_{n,i}\right)}\\
\bm{P}[n,co,ci] &= \\%\sum_{i,j=0,\cdots K-1} \bm{\bar{W}}[co,ci,i,j] \bm{\bar{A}}[n,ci,i,j]\\
            %&\quad=  \sum_{i,j=0,\cdots K-1} \left({\bm{Frac^{(w)}}}[co,ci,i,j] 2^{{\bm{Exp^{(w)}}}[co,ci,i,j]}\right)\times 
            %\left({\bm{Frac^{(a)}}}[n,ci,i,j] 2^{{\bm{Exp^{(a)}}}[n,ci,i,j]}\right)\\
            %&\quad=
           \sum_{i,j=0}^{K-1} & \left({\bm{Frac^{(w)}}}[co,ci,i,j]{\bm{Frac^{(a)}}}[n,ci,i,j]\right) \\
            & \times 2^{\left({\bm{Exp^{(w)}}}[co,ci,i,j] + {\bm{Exp^{(a)}}}[n,ci,i,j]\right)}
\end{split}
\end{medsize}
\end{equation}
where $\bm{Frac}$, $\bm{Exp}$ %are fractions and exponents, whose precision is 
are $(M_x+1)$-bit and $E_x$-bit. 

The intra-group calculation contains the multiplication of two $(M_x+1)$-bit values
%So the intro group MAC contains $(M_x+1)$-bit multiplication,
and $2\times(2^{E_x}-2)$-bit shifting. The resulting $(2M_x+2^{E_x+1}-2)$-bit integer values need to be accumulated with enough bit-width to get the partial sum $\bm{P}$. In previous 8-bit floating-point frameworks~\citep{fp8,hfp8}, the accumulator has to be floating-point since they use $E_x=5$. In contrast, we can use a 32-bit integer accumulator, since we adopt $E_x=2, M_x=4$ in the MLS tensor format on ImageNet. See Sec.~\ref{sec:analysis_accum} for more detailed analysis. \\ %, as shown in Table~\ref{tab:imagenet}.
%Since the accumulator might be float if $2M_x+2^{E_x+1}-2$ is too large, we need to reduce this bit-width to avoid using a costly floating-point accumulator~\citep{fullint}. \todo{mention reduce this bit width here?}
%because integer accumulator has much less cost than float one\citep{fullint}. The comparison is in Section~\ref{sec:exp}. 

\begin{table*}[bt]
  \centering
  \caption{Comparison of low-bit training methods on CIFAR-10 and ImageNet.
  Single number in the bit-width stands for fixed-point format bit-width, which is equivalent to $M_x$ and the corresponding $E_x$ is 0. ``f x'' indicates that x-bit floating-point numbers are used. ``ACCUM'' in the ``Bit-Width'' column stands for ``Accumulation'', while ``Acc.'' stands for ``Accuracy''.} %\zksays{We need to add more reference results(Dillon, Koester, Gysel, and Wang), and add more explain of WAEAcc and grouping method, and think about split to two tables.}}
   %If there is only one number in each bit-width description, it is $M_x$, and the description is equivalent to $\langle 0, M_x \rangle$. }%If there are two numbers of bit-width, they are $\langle E_x,M_x\rangle$, and it's $M_x$ if there is only one, which represents fixed-point quantization.} 
  %ome baselines of ResNet-18 are reproduced by their authors and we do the same.}
  \vskip 0.2in
  %\resizebox{\linewidth}{!}{
    \begin{tabular}{c|cccccc}
    \toprule
    Dataset & Method  & Model & Bit-Width (W/A/E/ACCUM) & Acc. & FP baseline & Acc. Drop\\
    \midrule
    \multirow{8}{*}{CIFAR-10}   & S2fFP8\citep{s2fp8}   & ResNet-20  & $\langle 5,2 \rangle$ $\langle 5,2 \rangle$ $\langle 5,2 \rangle$ f32  & 91.1\%  & 91.5\% & 0.4\%\\
                                & WAGE\citep{wage}    & VGG-like   & 2 8 8 32    & 93.2\%   & 94.1\%  & 0.9\%\\
                                & RangeBN\citep{rangebn} & ResNet-20  & 1 1 2 -     & 81.5\%   & 90.36\% & 8.86\%\\
                        %      & \citep{flexpoint}   & ResNet-110  & 16 16 16 32 & 94.8\%(t-5) & 95.2\%(t-5) & 0.4\%\\  
    \cmidrule(lr){2-7}
     & \multirow{6}{*}{Ours}  & ResNet-20       & 4 4 4 16   & 92.32\%     & 92.45\% & 0.13\%\\
     &                        & ResNet-20       & 2 2 2 16   & 90.39\%     & 92.45\% & 2.06\%\\
     &          & ResNet-20 & $\langle 2,1 \rangle$ $\langle 2,1 \rangle$ $\langle 2,1 \rangle$ 16 & 91.97\% & 92.45\% & 0.48\%\\
     &          & GoogleNet & $\langle 2,1 \rangle$ $\langle 2,1 \rangle$ $\langle 2,1 \rangle$ 16 & 93.95\% & 94.50\% & 0.55\%\\
     &          & VGG-16    & $\langle 2,1 \rangle$ $\langle 2,1 \rangle$ $\langle 2,1 \rangle$ 16 & 93.34\% & 93.76\% & 0.42\%\\
     &          & VGG-16    & $\langle 1,1 \rangle$ $\langle 1,1 \rangle$ $\langle 1,1 \rangle$ 8 & 92.77\% & 93.76\% & 0.99\%\\
    \hline
    \multirow{16}{*}{ImageNet} & FlexPoint\citep{flexpoint}& AlexNet  & 16 16 16 32 & 80.1\% (Top5) & 79.9\% (Top5) & -0.2\%\\
                               & DFP16\citep{dfp16}    & VGG-16    & 16 16 16 32  & 68.2\%  & 68.1\% & -0.1\%\\
                               & DFP16\citep{dfp16}    & GoogleNet & 16 16 16 32  & 69.3\%  & 69.3\% & 0\\
                               & RangeBN\citep{rangebn}  & ResNet-18 & 8 8 16 f32 & 66.4\% & 67.0\% & 0.6\%\\
                               & DoReFa\citep{dorefa}   & AlexNet   & 8 8 8 32  & 53.0\%  & 55.9\% & 2.9\%\\
                               & FullINT\citep{fullint}  & ResNet-18 & 8 8 8 32  & 64.8\%  & 68.7\% & 3.9\%\\
                               & FullINT\citep{fullint}  & ResNet-34 & 8 8 8 32  & 67.6\%  & 72.0\% & 4.4\%\\
                               & WAGE\citep{wage}     & AlexNet   & 2 8 8 32  & 48.4\%  & 56.0\% & 7.6\%\\
    & HFP8\citep{hfp8}    & ResNet-18 &$\langle5,3\rangle$ $\langle5,3\rangle$ $\langle5,3\rangle$ f32 & 69.0\% & 69.3\% & 0.3\%\\
    & S2FP8\citep{s2fp8}  & ResNet-18 &$\langle5,2\rangle$ $\langle5,2\rangle$ $\langle5,2\rangle$ f32 & 69.6\% & 70.3\% & 0.7\%\\
    & Ultra-Low\citep{ultra}  & ResNet-18 &4 4 $\langle3,1\rangle$ f16 & 68.3\% & 69.4\% & 1.1\%\\
    \cmidrule(lr){2-7}
    & \multirow{7}{*}{Ours}  & ResNet-18 & 8 8 8 32 & 68.5\% & 69.1\% & 0.6\%\\
    &                         & ResNet-18 & 4 4 4 16 & 66.5\% & 69.1\% & 2.6\%\\
    %&   & ResNet-18 & $\langle 1,6 \rangle$ $\langle 1,6 \rangle$ $\langle 1,6 \rangle$ 16 & 68.0\% & 69.1\% & 1.1\%\\
    &   & ResNet-18 & $\langle 2,4 \rangle$ $\langle 2,4\rangle$ $\langle 2,4 \rangle$ 32 & 68.2\% & 69.1\% & 0.9\%\\
    &   & ResNet-34 & $\langle 2,4 \rangle$ $\langle 2,4\rangle$ $\langle 2,4 \rangle$ 32 & 75.3\% & 76.1\% & 0.8\%\\
    &   & VGG-16    & $\langle 2,4 \rangle$ $\langle 2,4\rangle$ $\langle 2,4 \rangle$ 32 & 70.8\% & 70.9\% & 0.1\%\\
    &   & GoogleNet & $\langle 2,4 \rangle$ $\langle 2,4\rangle$ $\langle 2,4 \rangle$ 32 & 69.6\% & 69.5\% & -0.1\%\\
    \bottomrule
    \end{tabular}
    %}%
  \label{tab:imagenet}%
\end{table*}%

\noindent\textbf{Inter-group MACs} As for the inter-group calculation, each element in $\bm{S^{(p)}}$ is a $\langle E, 2\rangle$ number obtained by multiplying two $\langle E, 1\rangle$ numbers. % and it is , 
So it can be calculated by shift (multiplying the power of two) and addition as:
\begin{equation} 
\begin{split}
&\bm{Z}[co,x] = S_t^{(z)}\sum_{ci=0}^{C-1}{\bm{S^{(p)}}}[co,ci]\bm{P}[x,ci] =S_t^{(z)}\sum_{ci=0}^{C-1}\\ %S_t^{(z)} \\
            &\begin{cases}            
               \bm{P}[x,ci] 2^{-\bm{Exp^{(p)}}[co, ci]}                                              \\ 
               \bm{P}[x,ci] 2^{-\bm{Exp^{(p)}}[co, ci]}+\bm{P}[{x,ci}] 2^{-\bm{Exp^{(p)}}[co, ci]-1} \\
               \bm{P}[{x,ci}] 2^{1-\bm{Exp^{(p)}}[co, ci]}+\bm{P}[x,ci] 2^{-\bm{Exp^{(p)}}[co, ci]-2} 
            \end{cases}
            \label{eq:groupwise_scale}
\end{split}
\end{equation}
where the three cases correspond to $\bm{Man^{(p)}}[co,ci]$=00, $\bm{Man^{(p)}}[co,ci]$=01/10, and $\bm{Man^{(p)}}[co,ci]$=11, respectively. The index $n$ is ommited for simplicity and $x$ is used to denote the original 2-dimension spatial indexes $x, y$. \\

\noindent\textbf{Summarize of the convolution energy efficiency of the MLS format}
In the MLS format, \textit{the element-wise exponent} is 2-bit instead of 5-bit, thus the \textbf{intra-group accumulation} is simplified to use 32-bit integers. On the other hand, due to the special format of \textit{group-wise scaling factor}, $\bm{S^{(p)}}$ has a simple format, and the \textbf{inter-group accumulation} to calculate $\bm{Z}$ can be implemented efficiently on hardware without floating-point multiplication. Finally, the multiplication with the \textit{tensor-wise floating-point scaling factor} $S_t^{(z)}$ in Eq.~\ref{eq:groupwise_scale} can usually be omitted: $S_t^{(z)}$ only needs to be multiplied with the tensor-wise floating-point scale in the following layer instead of the feature map, as long as there is no following element-wise addition on $Z$ with another tensor. %(e.g., skip connections).

\subsection{Analysis of Accumulation Bit-Width}
\label{sec:analysis_accum}
Convolution consists of multiplication and accumulation. When different data formats are used, the results of multiplication have different dynamic ranges. As specified by the IEEE 754 standard, the gradual underflow behavior of a floating-point representation that has  M-bit mantissa ($Man$) and E-bit exponent ($Exp$) is as follows.
%When the input data of convolution has, 
If $Exp$ is not equal to the minimum value, the float value is not underflow, and is calculated as 
\begin{equation}
\begin{split}
    float &= Frac \times 2^{-Exp} \\
          &=\left(1+\frac{Man}{2^M}\right)\times 2^{-Exp}.
\end{split}
    \label{equ:float}
\end{equation} 

If $Exp$ is equal to the minimum value, the float value is an gradual-underflowed value, and is calculated as 
\begin{equation}
\begin{split}
    float &= Frac \times 2^{-Exp} \\
          &=\left(0+\frac{Man}{2^M}\right)\times 2^{-Exp},
\end{split}
    \label{equ:float-underflow}
\end{equation} 
where $Frac$ is $(M+1)$-bit fraction, calculated by adding a 0 or 1 at the highest bit of mantissa. 

%As discussed in Sec. 4.2 in the paper, we calculate the product of two numbers as
The product of two numbers is calculated as
\begin{equation}
\begin{split}
    float_1 \times float_2 &= Frac_1 \times 2^{-Exp_1} \times Frac_2 \times 2^{-Exp_1}\\
          &= \left( Frac_1 \times Frac_2 \right) \times 2^{-Exp_1-Exp_2},
\end{split}
    \label{equ:multiple}
\end{equation} 
where $Frac_1 \times Frac_2$ is a $(M+1)$-bit multiplication, and the result is $(2M+2)$-bit. 
Since the minimum value of exponent is used to represent underflow, E-bit $Exp$ represents $2^{E}-1$ levels and ``$\times 2^{-Exp}$'' is $(2^{E}-2)$-bit shifting. Therefore, ``$\times 2^{-Exp_1-Exp_2}$'' is $(2^{E+1}-4)$-bit shifting, and the final result of floating-point multiplication has a dynamic range of $2M+2+2^{E+1}-4 = (2M+2^{E}-2)$-bit. These resulting $(2M+2^{E+1}-2)$-bit integer values need to be accumulated with enough bit-width to get the partial sum. In previous 8-bit floating-point frameworks, the accumulator has to be floating-point since they use $E=5$. In contrast, we can use a 32-bit integer accumulator, since we adopt $E=2, M=4, (2M+2^{E+1}-2)=14$ in the MLS tensor format on ImageNet.

\section{Experiments}
\label{sec:exp}
%\subsection{Experimental Setup}
%, leaving the first and the last layer unquantized%following previous studies
%~\citep{dorefa,mixedprecision,hfp8}. %, but all downsample conlolutions are quantized.
% In all the experiments,
% %bellow, 
% we use the standard data augmentation and hyper-parameters~\citep{resnet}. For both datasets, SGD with momentum 0.9 and weight decay 5e-4 is used, and the initial learning rate is set to 0.1. We train the models for 90 epochs on ImageNet, and decay the learning rate by 10 every 30 epochs. On CIFAR-10, we train the models for 160 epochs and decay the learning rate by 10 at epoch 80 and 120.

\subsection{Experimental Setup} 
We train ResNet~\citep{resnet}, VGG~\citep{vgg}, and GoogleNet~\citep{googlenet} on CIFAR-10~\citep{cifar} and ImageNet~\citep{imagenet} with our low-bit training framework.
In all the experiments, the first and the last layer are left unquantized following previous studies~\citep{dorefa,mixedprecision,hfp8}.
On both CIFAR-10 and ImageNet, SGD with momentum 0.9 and weight decay 5e-4 is used, and the initial learning rate is set to 0.1. We train the models for 90 epochs on ImageNet, and decay the learning rate by 10 every 30 epochs. On CIFAR-10, we train the models for 160 epochs and decay the learning rate by 10 at epoch 80 and 120. We experiment with the MLS tensor formats using different $\langle E_x, M_x\rangle$ configurations, the group-wise scaling are in $\langle8,1\rangle$ format for all experiments in Table~\ref{tab:imagenet}. And we adopt the same quantization bit-width for weight, activation and error for a simpler hardware design.

\subsection{Results on CIFAR-10 and ImageNet} 
The training results on CIFAR-10 and ImageNet are shown in Table~\ref{tab:imagenet}. We can see that our method can achieve a better balance between higher accuracy and lower bit-width.
%The results of training ResNet-20 on CIFAR-10 
% are shown in Table~\ref{tab:cifar}.
% 
Previous study~\citep{dorefa} found that quantizing error to a low bit-width hurt the accuracy, but our method can quantize error to $M_x=1$ on CIFAR-10, with a small accuracy drop of 0.48\%, 0.55\%, and 0.42\% for ResNet-20, GoogleNet, and VGG-16, respectively. %from 92.45\% to 91.97\%.
%We also apply our framework to train ResNet-18 on ImageNet, and Table~\ref{tab:imagenet} shows the comparison of our method and other low-bit training methods.

On ImageNet, the accuracy degradation of our method is rather minor under 8-bit quantization (0.6\% accuracy drop from 69.1\% to 68.5\%), which is comparable with other state-of-the-art work. 
In the cases with lower bit-width, our method achieves a higher accuracy (66.5\%) with only 4-bit than ~\citet{rangebn} who uses 8-bit (66.4\%). 
With $\langle 2, 4 \rangle$ data format, 
for all the models including ResNet-18, ResNet-34, VGG-16, and GoogleNet, our method can achieve an accuracy loss less than 1\%.
In this case, the bit-width of the intermediate results
%needed to be accumulated % in $P$ calculation 
is $2M_x+2^{E_x+1}-2=14$, 
which means that the accumulation can be conducted using integers, instead of floating-points~\citep{hfp8,s2fp8}, as we disscussed in Sec.~\ref{sec:mls-conv}. Although a previous work~\citep{ultra} quantizes W/A/E to 4-bit, the three different types of Convs between them are different, which requires three different unit implementations. In contrast, our work unifies the W/A/E format and the Conv calculation, thus requires only one type of Conv unit.

\begin{table}[tb]
    \centering
    \caption{Number of operations and sensitivity of ResNets, VGG-16, and GoogleNet.}
    \label{tab:model_flops}
    \begin{tabular}{c|cccc}
        \toprule
      \multirow{2}{*}{Model} & \multirow{2}{*}{Inference GOPs} & \multirow{2}{*}{\begin{tabular}[c]{@{}c@{}} Acc. Drop of\\ 6-bit Training\end{tabular}}\\
       & & \\\midrule
       ResNet-18  &	1.88  & 0.9\%\\
       ResNet-34	 &  3.59  & 0.8\%\\
       VGG-16     &  15.25 & 0.1\%\\
       GoogleNet &  1.58  & -0.1\%\\
         \bottomrule
    \end{tabular}
\end{table}

%\new{Analyze which types of NN architectures are more insensitive to quantization: }
%To summarize, our method can simplify the accumulation compared with previous 8-bit floating-point training methods, while retaining a similar accuracy with full precision training.
%With the help of the 2-bit element-wise exponent, training with only 4-bit mantissa can achieve an accuracy degradation less than 1\%. 
%which is much less than  other floating-point training methods.
%It is worth noting
We note that the performance of VGG and GoogleNet CNN models in low-bit training is better than ResNets. 
%We think that this is because the number of channels of ResNet is slightly less than VGG and GoogleNet when the network depth configuration is similar, which reduces the redundancy of expression ability and is more sensitive to quantization error.
We think that this is because there are fewer channels in ResNet than VGG and GoogleNet when the network depth configuration is similar. And the smaller redundancy of ResNets makes them more sensitive to quantization errors.
In fact, VGG-16 is 7 times as much as ResNet-18 in terms of computation, as shown in Table~\ref{tab:model_flops}. That means even if ResNet adopts a higher bit-width and higher accuracy scheme for training, it still has a higher energy efficiency. In contrast, the model structure of GoogleNet class shows a higher adaptability in the face of low-bit training scenarios, which brings inspiration to the future network architecutre design for low-bit training scenarios. %\todo{ Give out the FLOPs number of ResNet and VGG17, GoogleNet?}

\subsection{Ablation Studies}
%Using a tensor-wise scaling factor is a common practice~\citep{flexpoint,gemmlowp}, so we  use tensor-wise scaling in various experiments bellow, and we will check the effects of group-wise scaling and element-wise exponent. 

\begin{figure*}[ht]
\vskip 0.2in
\begin{center}
% \centerline{\includegraphics[width=\linewidth]{figures/errors_cifar.pdf}}
\centerline{\includegraphics[width=\linewidth]{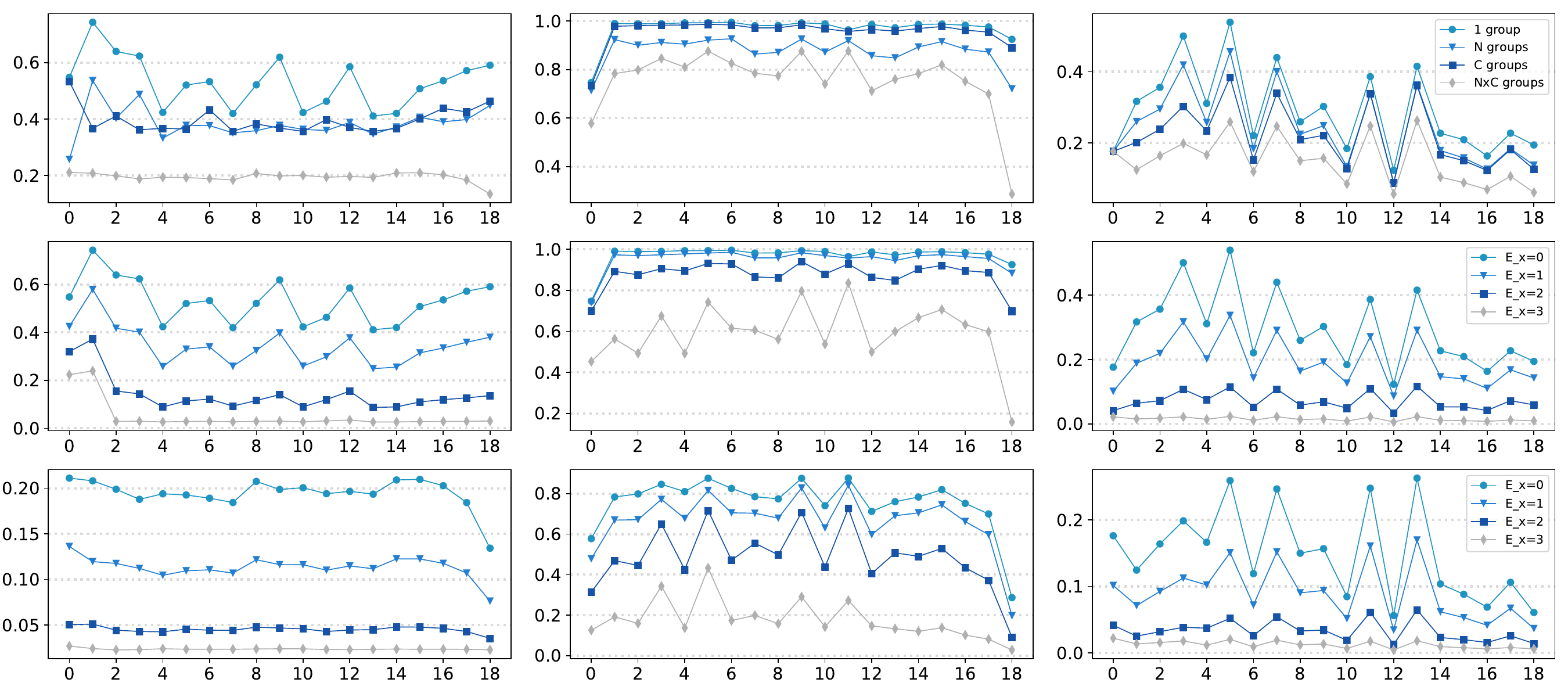}}
\caption{Average relative quantization errors (AREs) of weight, error, activation (left, middle, right) in each layer when training a ResNet-20 on CIFAR-10. X axis: Layer index. Row 1: Different grouping dimensions ($\langle 0, 3\rangle$ formatted $\bar{\bm{X}}$, $\langle 8,1\rangle$ formatted $\bm{S_g}$); Row 2: Different $E_x$ ($\langle E_x, 3\rangle$ formatted $\bar{\bm{X}}$, no group-wise scaling);
Row 3: Different $E_x$ ($\langle E_x, 3\rangle$ formatted $\bar{\bm{X}}$, $\langle 8, 1\rangle$ formatted $\bm{S_g}$, $N\times C$ groups).}
\label{pic:errors}
\end{center}
\vskip -0.2in
\end{figure*}

\begin{table}[tb]
    \centering
    \caption{Accuracy of training ResNet-20 on CIFAR-10. ``Div.'' means that the training failed to converge. ``None'' means that group-wise scaling is not used (\#group=1).} 
    \label{tab:ablation}
    \begin{small}
    \begin{tabular}{ccc|ccccccc}
        \toprule
        \#group & $M_g$ & $E_x$ & $M_x$=4 & $M_x$=3 & $M_x$=2 & $M_x$=1 &  \\
        \midrule
        1       &None  &  0 & 90.02  & 85.68  & Div.   & Div.      &  \\
        c       & 0    &  0 & 91.54  & 88.35  & 82.29  & Div.      &  \\
        n       & 0    &  0 & 91.78  & 89.62  & 80.71  & Div.      &  \\
        nc      & 0    &  0 & 92.14  & 91.64  & 88.97  & 76.98  & \\
        nc      & 1    &  0 & 92.37  & 91.73  & 90.39  & 82.61  &\\
        \midrule
        1       &None    & 0 &  90.02  & 85.68  & Div.   & Div.   & \\
        1       &None    & 1 & 91.67  & 90.11  & 84.72  & 70.4   & \\
        1       &None    & 2 & 92.32  & 92.34  & 91.58  & 90.32  &\\
        %1       &None    & 3 & 92.66  & 92.41  & 92.47  & 92.04  &  \\
        \midrule
        nc      &1   & 0    &   92.37  & 91.73  & 90.39  & 82.61  &  \\
        nc      &1   & 1    &   92.52  & 92.16  & 91.48  & 89.97  &\\
        nc      &1   & 2    &   92.37  & 92.65  & 92.05  & 91.97  & \\
        \bottomrule
    \end{tabular}
    \end{small}
\end{table}

\subsubsection{Group-wise Scaling}%Group-wise Scaling}
%In this paper, we denote the filter coefficient and feature map of convolution as weight (W) and activation (A), respectively. In the back-propagation, the gradient of convolution results and weights are denoted as error (E) and gradient (G), respectively. %W, A, and E of each layer are 4-dimension tensors in training. 
%For A and E, the four dimensions are sample in batch (N) , channel (C) , feature map height (H) , and feature map width (W) , respectively.
%the range of data various in different groups. 

Group-wise scaling is beneficial as the data ranges vary across different groups. We compare the average relative quantization errors (AREs) of using the three grouping dimensions (Sec.~\ref{sec:mls_mapping}) with $\langle 8,1\rangle$ group-wise scaling format and $\langle 0,3\rangle$ element format. %on real data recorded from ResNet-20 training on CIFAR-10 and take the average relative quantization error (ARE) as the measurement.
The first row of Fig.~\ref{pic:errors} shows that 
%It is shown in the first row of Fig.~\ref{pic:errors} that 
the AREs are smaller when each tensor is split to $N\times C$ groups.
Furthermore, we compare these grouping dimensions in the training process. The first section of Table~\ref{tab:ablation} shows that when tensors are split to $N\times C$ groups, the training accuracy is higher. This indicates that the reduction of AREs is important for the accuracy of low-bit training.  % thaen that when tensors are split to $N$ or $ C$ or $1$ group(s).
%And we can see that the $\langle E_g, 1\rangle$ format group-wise scaling plays a important role, .
And we can see that $M_g=1$ is important for the accuracy, especially with low $M_x$ (e.g., when $M_x$=1, $76.98\%$ V.S. $82.61\%$).

%\new{Add a paragraph and a table column on error-only group-wise ablation study. }
To show that the low-bit training is distinct from previous efficient inference studies, we add an ablation study of ``error'' format to further demonstrate: W/A are quantized with group-wise scaling factors consists of only exponent, when error is also quantized as this, the accuracy is $90.7\%$. After introducing our $\langle8, 1\rangle$ group-wise scaling for error, the accuracy is $91.9\%$. This indicates that MLS format is better than shared exponent in \citep{msfp} for a benign training.
%As discussed in Section~\ref{sec:method-groupwise}, independent group-wise high-precision scaling factors could not be used in hardware-friendly quantization. The simplest way is to use the power of 2 as the group-wise scales. Besides this simple solution, we propose another two group-wise scaling methods: ``same mantissa scale'' and ``simple mantissa scale''. 
%As shown in Table~\ref{tab:group-scale}, the group-wise scaling methods could achieve significantly better results than using the floating-point scale in the cases of low bit-width (e.g., above 90.0 \% V.S. 78.95 \%). \ztcsays{weird next sentence, maybe "we found out that simple mantissa strikr a better balance betwwen computation and accuracy"} Among these group-wise scaling methods, ``simple mantissa scale'' group-wise scaling is the best.

\subsubsection{Element-wise Exponent}
To study the influence of the element-wise exponent, we compare the AREs of quantization with different $E_x$ without group-wise scaling, and the results are shown in the second row of Fig.~\ref{pic:errors}. Intuitively, using more exponent bits results in larger dynamic ranges and smaller AREs. And with larger $E_x$, the AREs of different layers are closer. %\todo{?what reason} 
% Besides the ARE evaluation, we also show the training accuracies with different $\langle E_x, M_x \rangle$ configurations in Table~\ref{tab:ablation}.
% %Fig.~\ref{pic:different-exp}. 
% We can see that using larger $E_x$ achieves better accuracies, %across different $M_x$ choices
% especially when $M_x$ is extremely small.
Besides the ARE evaluation, Table~\ref{tab:ablation} shows that a larger $E_x$ achieves a better accuracy, %across different $M_x$ choices
especially when $M_x$ is extremely small.

%The accuracy of low-bit training with different $\langle E_x, M_x\rangle$ configurations without group-wise scaling is shown in Fig.~\ref{pic:different-exp}, training with more $E_x$ achieves better performance across different $M_x$ choices, especially in the cases with an extremely low bit-width.

As shown in Fig.~\ref{pic:errors} Row 3 and Table~\ref{tab:ablation}, when jointly using the group-wise scaling and the element-wise exponent, the ARE and accuracy are further improved.
And we can see that the group-wise scaling is important for simplifying the floating-point accumulator to a fixed-point one: One can use a small element-wise exponent with group-wise scaling  (\#group=nc, $M_g$=1, $E_x$=0,  Acc.=$92.37\%$) to get a comparable accuracy to a configuration with larger $E_x$=2 without group-wise scaling (Acc.=$92.32\%$).
%When using the group-wise scaling together with the element-wise exponent, the AREs are smaller than using only one of them, as shown in Fig.~\ref{pic:errors}. The low-bit training results of the combination of the two scalings is shown in %Fig.~\ref{pic:combination}
%Table~\ref{tab:ablation}, \todo{seems next sentence 's result is obvious? maybe delete it} and they are consistently better than the results of only using one scaling across different $M_x$ choices. And we can see that the group-wise scaling reduces the $E_x$ needed to achieve the same precision, and that is very important to simplify the floating-point unit to the fixed-point unit.

\subsection{Hardware Energy Consumption}

%Our method simplifies the convolution, which account for the main computational cost of CNN training. 
Fig.~\ref{pic:hardware-arch} shows a typical convolution hardware architecture, which consists of three main components: local multiplication (MUL), local accumulation (LocalACC), and addition tree (TreeAdd). Our framework mainly improves the local multiplications and accumulations. Compared with the full-precision design, we simplify the
floating-point multiplication (FP MUL) to use a bit-width less than 8 and the local floating-point (FP ACC) to use 16-bit or 32-bit integer. 
%According to the data reported by \citep{fullint}, the energy efficiencies of the integer design are at least 7$\times$ and 20$\times$ higher than full-precision one, respectively.
To evaluate the energy consumption, we implement the RTL design of the MAC unit with different arithmetic. Table~\ref{tab:evaluation} shows the hardware power results given by Design Compiler simulation with TSMC 65nm process and 1GHz clock frequency.
Then, using the numbers of different operations in convolution, we can estimate our energy efficiency improvement ratio $r$ in a single $3\times3$ convolution as
%Taking the training process of ResNet-18 on ImageNet as an example, the energy efficiency improvement ratio is 
\begin{equation}
\begin{medsize}
\begin{split}
    r &= \left[ 2.311(\#MUL) + 0.512(\#LocalACC) + 0.512(\#TreeAdd)\right] \\
    &\div [0.124(\#MUL) + 0.065(\#LocalACC+\#GroupwiseScale)\\
    &+ 0.512(\#TreeAdd)] \approx 11.5
\label{equ:approximation}
\end{split}
\end{medsize}
\end{equation}
where $GroupwiseScale$ is the group-wise scaling that could be implemented efficiently as in Eq.~\ref{eq:groupwise_scale}. The energy estimation of $3\times3$ convolutions is also shown in Fig.~\ref{fig:energy-illustration}. Next, in Sec.~\ref{sec:energy_estimation_details}, we present the energy analysis details of the whole network, which takes the energy consumption of different types of operations into consideration.
%The detailed energy estimation of different types of operations are shown in the appendix. %Table~\ref{tab:detailed-energy}.

\begin{table}[tb]
    \centering
    \caption{The power evaluation (mW) results of MAC units with different arithmetic, simulated by Design Compiler with TSMC 65nm process and 1GHz clock.}
    \label{tab:evaluation}
    \begin{tabular}{c|cc}
        \toprule
        Operation & MUL & LocalAcc \\
        \midrule
        Full Precision                      & 2.311   & 0.512       \\
        8-bit FP~\citep{hfp8}               & 0.105   & 0.512       \\
        8-bit INT~\citep{fullint}           & 0.155   & 0.065       \\
        Ours                                & 0.124   & 0.065       \\
        \bottomrule
    \end{tabular}
\end{table}

\begin{table*}[tb]
\centering
\caption{The comparison of the detailed energy estimation of training ResNet-34 on ImageNet using full-precision training and our low-bit training framework. ``DQ'' means dynamic quantization, which is an additional operation in our framework.}
\begin{tabular}{c|cccccc}
\toprule
\multirow{2}{*}{Op Name}    & \multicolumn{3}{c}{Full Precision Training}     & \multicolumn{3}{c}{Our Low-Bit Training}    \\\cmidrule(lr){2-4}\cmidrule(lr){5-7}
                            & {Op Type}  & { Op Amount} & { Energy/$\mu$J}    & { Op Type}  & { Op Amount}         & { Energy/$\mu$J}  \\
\midrule
\multirow{3}{*}{Conv}       & { FloatMul} & { 1.12E+10}  & { 25900}           & { FP7Mul}   & { 1.12E+10}          & { 1390} \\
                            & { FloatAdd} & { 1.12E+10}  & { 5740}            & { IntAdd}   & { 1.12E+10}          & { 729} \\
                            & {- }         & {0 }          & {0 }             & { FloatAdd} & { 1.21E+09}          & { 620} \\
\midrule
\multirow{2}{*}{BN}         & { FloatMul} & { 4.87E+07}  & { 101}             & { FloatMul}  & { 4.87E+07}          & { 101}  \\
                            & { FloatAdd} & { 4.38E+07}  & { 24.9}            & { FloatAdd}  & { 4.38E+07}          & { 24.9} \\
\midrule
\multirow{2}{*}{FC}         & { FloatMul} & { 3.07E+06}  & { 7.1}             & { FloatMul}  & { 3.07E+06}          & { 7.1}  \\
                            & { FloatAdd} & { 3.07E+06}  & { 1.57}            & { FloatAdd}  & { 3.07E+06}          & { 1.57} \\
\midrule
\multirow{2}{*}{SGD Update} & { FloatMul} & { 5.16E+07}  & { 119}             & { FloatMul}  & { 5.16E+07}          & { 119}  \\
                            & { FloatAdd} & { 5.16E+07}  & { 26.4}            & { FloatAdd}  & { 5.16E+07}          & { 26.4} \\
\midrule
\multirow{2}{*}{DQ}         &  \multirow{2}{*}{-}       &  \multirow{2}{*}{0}           &  \multirow{2}{*}{0}                 & { FloatMul}  & { 3.90E+7 + 6.88E+7} & { 249} \\
                            &         &      &                 & { FloatAdd}  & { 1.95E+6 + 3.44E+7} & { 27.6} \\
\midrule
\multirow{2}{*}{EW-Add}     & { FloatAdd} & { 2.88E+06}  & { 1.47}            & { FloatAdd}  & { 2.88E+06}          & { 1.47} \\
                            & { -}         & { 0}          & { 0}                & { FloatMul}  & { 2.88E+06}          & { 6.66} \\

\midrule
Sum                         & { }         & { }          & { 32000}           & { }          & { }                  & { 3130}   \\
\bottomrule
\end{tabular}
\label{tab:detailed-energy}%
\end{table*}

\subsection{Energy Estimation Details}
\label{sec:energy_estimation_details}

For completeness, we give the detailed energy estimation of different operation types when training ResNet-34 on ImageNet in Table~\ref{tab:detailed-energy}, in which all overheads introduced by our method are considered. The energy consumption is calculated by multiplying the operation amount (Table~\ref{tab:ops}) and the energy consumption of each operation (Table~\ref{tab:evaluation}).

Considering a convolution with $C_i$ input channels, $C_o$ output channels, $K\times K$ kernel size, and $W\times H$ feature map size, the operation amounts of floating-point multiplications and additions are $C_i \times C_o \times K \times K \times W \times H$, and the operation amount in the whole network is calculated by accumulating the operation amounts of each layer in both the forward and backward processes. In our low-bit training framework, floating-point additions are only reserved in the adder tree, and the amount is $C_i \times C_o \times W \times H$. The amount of integer accumulation is equal to the other local addition and shifting (which is the same as the adder tree). The \textbf{group-wise scaling factors} introduce additional scaling. % between intra-group MACs and inter-group MACs. 
Fortunately, when using the $\langle E_g,0\rangle$ or $\langle E_g,1\rangle$  format, we can implement the group-wise scaling efficiently with shifting (Eq.~\ref{equ:group-scaling}). And the energy consumption is comparable to a LocalACC operation. We have already taken this overhead into account when estimating the energy efficiency improvement ratio of convolution in Eq.~\ref{equ:approximation}.

For \textbf{batch normalization}, fully connected layer, SGD update, the operation amount and energy consumption are the same for both the full-precision and our low-bit training framework. Specifically, 9 multiplications and 10 additions are performed on each element of a $C \times W\times H$ feature map in the forward and backward processes for batch normalization. The forward process of batch normalization is:
\begin{equation}
\begin{split}
    \mu      &= \frac{1}{M} \sum_{i=1}^{M} x_i \\
    \sigma^2 &= \frac{1}{M} \sum_{i=1}^{M} x_{i}^2 - \mu^2 \\
    y_i      &= \frac{x_i - \mu}{\sqrt{\sigma^2+0.00005}}\\
    z_i      &= \gamma y_i + \beta.
\end{split}
\end{equation}
We can see that in the forward process of batch normalization, for each input element, one addition is required to calculate the batch mean, and one multiplication and one addition are used to calculate the batch variance, and two multiplications and two additions are used for normalization and affine transformation. 

The backward process of batch normalization is:
\begin{equation}\label{eq:back_bn}
\begin{split}
    \frac{\partial L}{\partial \gamma} &= \sum_{i=1}^{M} \frac{\partial L}{\partial z_i} \cdot y_i \\
    \frac{\partial L}{\partial \beta}  &= \sum_{i=1}^{M} \frac{\partial L}{\partial z_i}\\
    \frac{\partial L}{\partial y_i}  &= \frac{\partial L}{\partial z_i} \cdot \gamma\\
    t_1 &= \sum_{j=1}^{M} \frac{\partial L}{\partial y_j} \\
    t_2 &= \sum_{j=1}^{M} (\frac{\partial L}{\partial y_j} \cdot y_j) \\
    \frac{\partial L}{\partial x_i}  &= \frac{M\frac{\partial L}{\partial y_i} - t_1 - y_i \cdot t_2}
    {M\cdot \sqrt{\sigma^2 + 0.00005}}.
\end{split}
\end{equation}
There are six multiplications and six additions performed on one element in the backward process of batch normalization (``1M1A, 1A, 1M, 1A, 1M1A, 3M2A'' for each formula in Eq.~\ref{eq:back_bn}, respectively). As shown in Table~\ref{tab:ops}, the number of multiplication and addition operation in batch normalization is orders of magnitude smaller than that in the convolutions. Hence, the energy consumption of batch normalization is relatively smaller compared with convolution.

As for \textbf{dynamic quantization}, we consider that 4 multiplications and 2 additions are needed for one element: one addition is to calculate the max (Alg.~\ref{alg:dynamic-quantization} Line 2 in the original paper) and the other one is to calculate the sum of $\bm{Frac_{xs}}$ and $\bm{R}$ (Alg.~\ref{alg:dynamic-quantization} Line 13 in the original paper), and the four multiplications are used for the Alg.~\ref{alg:dynamic-quantization} Line 4 and Line 9 in the original paper. Note that other multiplications and divisions in Alg.~\ref{alg:dynamic-quantization} describe the simulation of the dynamic quantization process on the floating-point platform, and they do not actually introduce overhead. The number of elements is $C \times W \times H$ for activation and error, and $C_i \times C_o \times K \times K$ for weight, and their energy consumption are shown separately in Table~\ref{tab:detailed-energy}.

For \textbf{element-wise addition} of two MLS tensors $z_1, z_2$, we need to multiply the ratio of their tensor-wise scales $S^{z_1}_t/S^{z_2}_t$ to $Z_2$, and then the element-wise addition can be conducted. Therefore, extra multiplications of the same amount are needed in our low-bit training framework. The last row of Table~\ref{tab:detailed-energy} shows the sum of the energy consumption of previous operations, and the results are not exactly the sum of the numbers in previous rows. The results show that our low-bit training framework achieves $32000\div 3130 = 10.2\times$ higher energy efficiency than full-precision training. The energy consumption calculation of other networks and 8-bit floating-point training can be conducted similarly as the above analysis, and is not discussed here.

%Since the energy consumption of floating-point multiplication is about 4 times of addition (Appendix Tab. 3)~\citep{fixfloat}, %
To summarize, the introduced overhead of our framework is low compared with the reduced cost. Taking all overheads into consideration, we can estimate that our whole low-bit training framework could achieve $8.3\sim10.2\times$ higher energy efficiency than the full-precision framework when training different models on ImageNet. Compared with previous low-bit floating-point training frameworks~\citep{hfp8}, our framework can achieve $1.9\sim2.3\times$ higher energy efficiency due to the simplified integer accumulator. 
 
 %The example of detailed energy estimates are shown in Table~\ref{tab:detailed-energy}.
%,, without considering the simplification of floating-point multiplication.
%\todo{hardware exp}

%According to the cost of different operations in Table~\ref{tab:fix-float}, our algorithm has the potential to achieve at least 4.6 $\times$ energy efficiency of the computation compared with full precision when training ResNet. Actually the internal memory cost will also be alleviated because the low-bit tensor format is used. 

\section{Conclusion}
\label{sec:conclusion}
This paper proposes a low-bit training framework to enable training CNNs with lower bit-width convolution while retaining the accuracy. Specifically, we design a multi-level scaling (MLS) tensor format containing tensor-wise scaling, group-wise scaling and element-wise scaling. And we describe the corresponding quantization procedure and low-bit convolution arithmetic, and analyze why our data format and hardware design bring energy efficiency improvements. In the hardware implementation, instead of using traditional systolic array hardware architecture, we adopt an adder tree architecture hardware to support our MLS data format. 
Experimental results and the energy consumption simulation of the corresponding computing unit demonstrate the effectiveness of our framework. Compared with previous low-bit integer training frameworks, our framework can retain a higher accuracy for a variety of models, including ResNets, VGG, and GoogleNet. Compared with previous low-bit floating-point training frameworks, our framework can achieve much higher energy efficiency. 
%And this work shows that adder tree architecture supporting group-wise scaling might be more suitable for CNN training than systolic array architecture.%\todo{Where?}

\appendices

\section*{Acknowledgment}

This work was supported by National Natural Science Foundation of China (No. U19B2019, 61832007, 61621091); Beijing National Research Center for Information Science and Technology (BNRist); Beijing Innovation Center for Future Chips; Beijing Academy of Artificial Intelligence. And we thank Huawei Technologies for the support and valuable discussion.

\ifCLASSOPTIONcaptionsoff
  \newpage
\fi



\bibliographystyle{IEEEtran}
\bibliography{IEEEabrv,refs}


% \end{thebibliography}


%\newpage

\vspace{-1.0cm}
\begin{IEEEbiography}[{\includegraphics[width=1.2in,height=1.15in,clip,keepaspectratio]{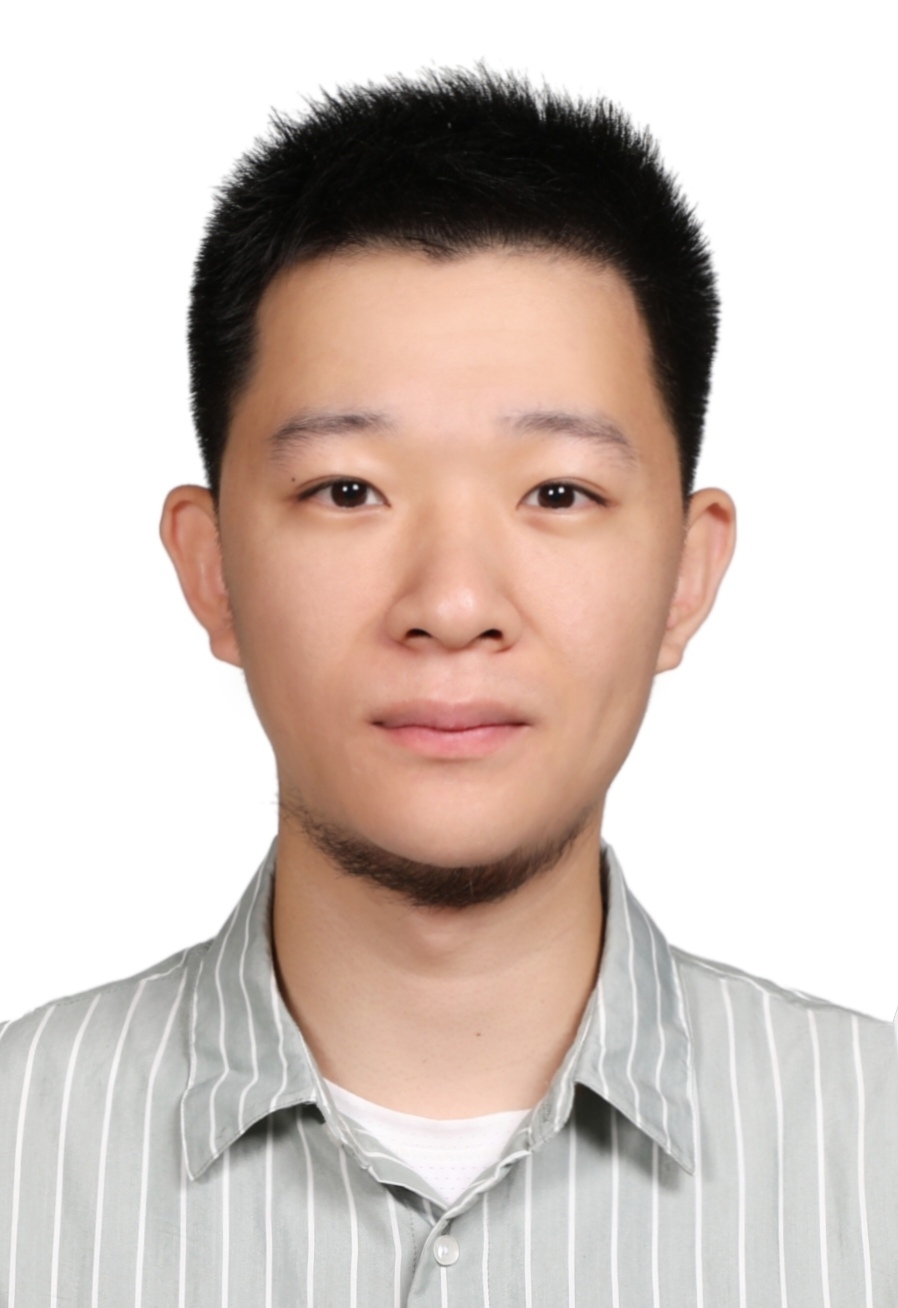}}]{Kai Zhong}
received his B.S. degree in electronic engineering from Tsinghua University, Beijing, in 2019. He is currently pursuing his Ph.D. degree at the Department of Electronic Engineering, Tsinghua University, Beijing. 
His research mainly focuses on deep learning acceleration and hardware-friendly algorithm optimization.
\end{IEEEbiography}

\vspace{-1.0cm}
\begin{IEEEbiography}[{\includegraphics[width=1.2in,height=1.25in,clip,keepaspectratio]{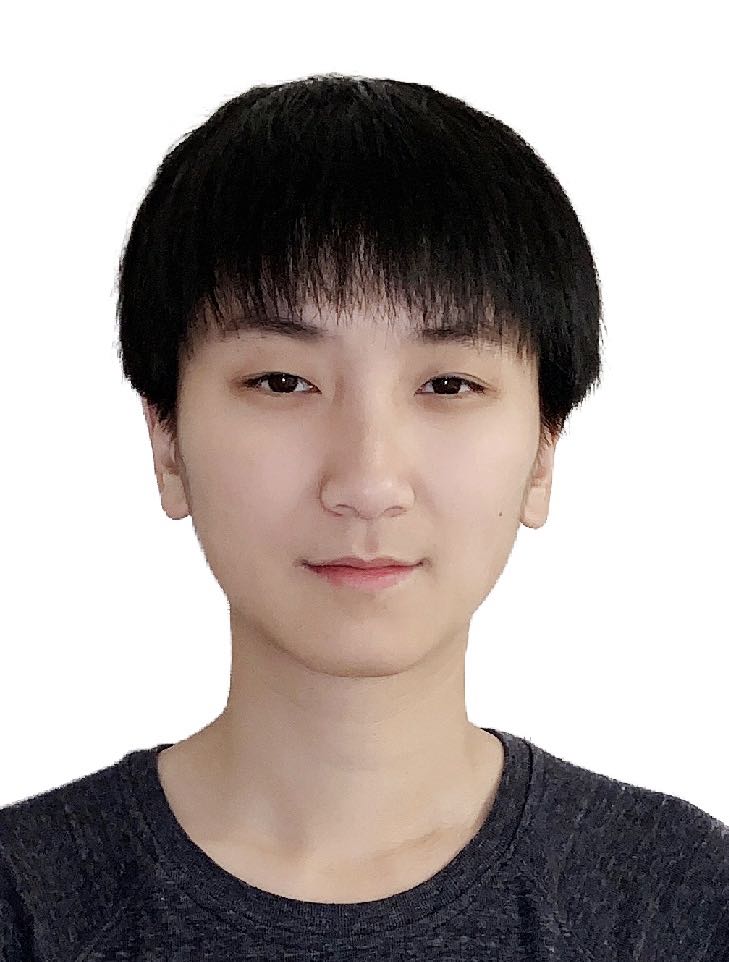}}]{Xuefei Ning} received her B.S. degree in electronic engineering from Tsinghua University in 2016, and is currently pursuing a Ph.D. degree at EE Department, Tsinghua University. Xuefei's research mainly focuses on efficient deep learning algorithm design and neural architecture search.
\end{IEEEbiography}

\vspace{-1.0cm}
\begin{IEEEbiography}[{\includegraphics[width=1in,height=1.25in,clip,keepaspectratio]{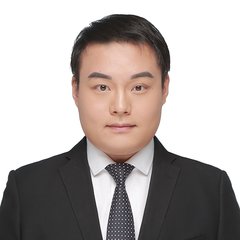}}]{Guohao Dai}(S’18) is currently a postdoctoral researcher in the Department of Electronic Engineering, Tsinghua University, Beijing, China. He received the B.S. degree and Ph.D. degree (with honor) in Department of Electronic Engineering from Tsinghua University in 2014 and 2019. Guohao’s research mainly focuses on graph computing, hardware virtualization, heterogeneous hardware computing, cloud computing, and etc. He has received Best Paper Award in ASPDAC 2019, and Best Paper Nomination in DATE 2018.
\end{IEEEbiography}

\vspace{-1.0cm}
\begin{IEEEbiography}[{\includegraphics[width=1in,height=1.25in,clip,keepaspectratio]{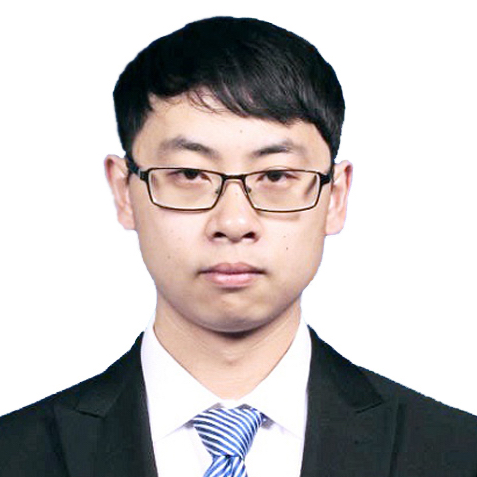}}]{Zhenhua Zhu}
received his B.S. degree in electronic engineering department of Tsinghua University, Beijing, China, in 2018. He is currently pursing his Ph.D degree in electronic engineering department of Tsinghua University. His research mainly focuses on memristor, computer architecture, and Processing-In-Memory.
\end{IEEEbiography}

\vspace{-1.0cm}
\begin{IEEEbiography}[{\includegraphics[width=1in,height=1.25in,clip,keepaspectratio]{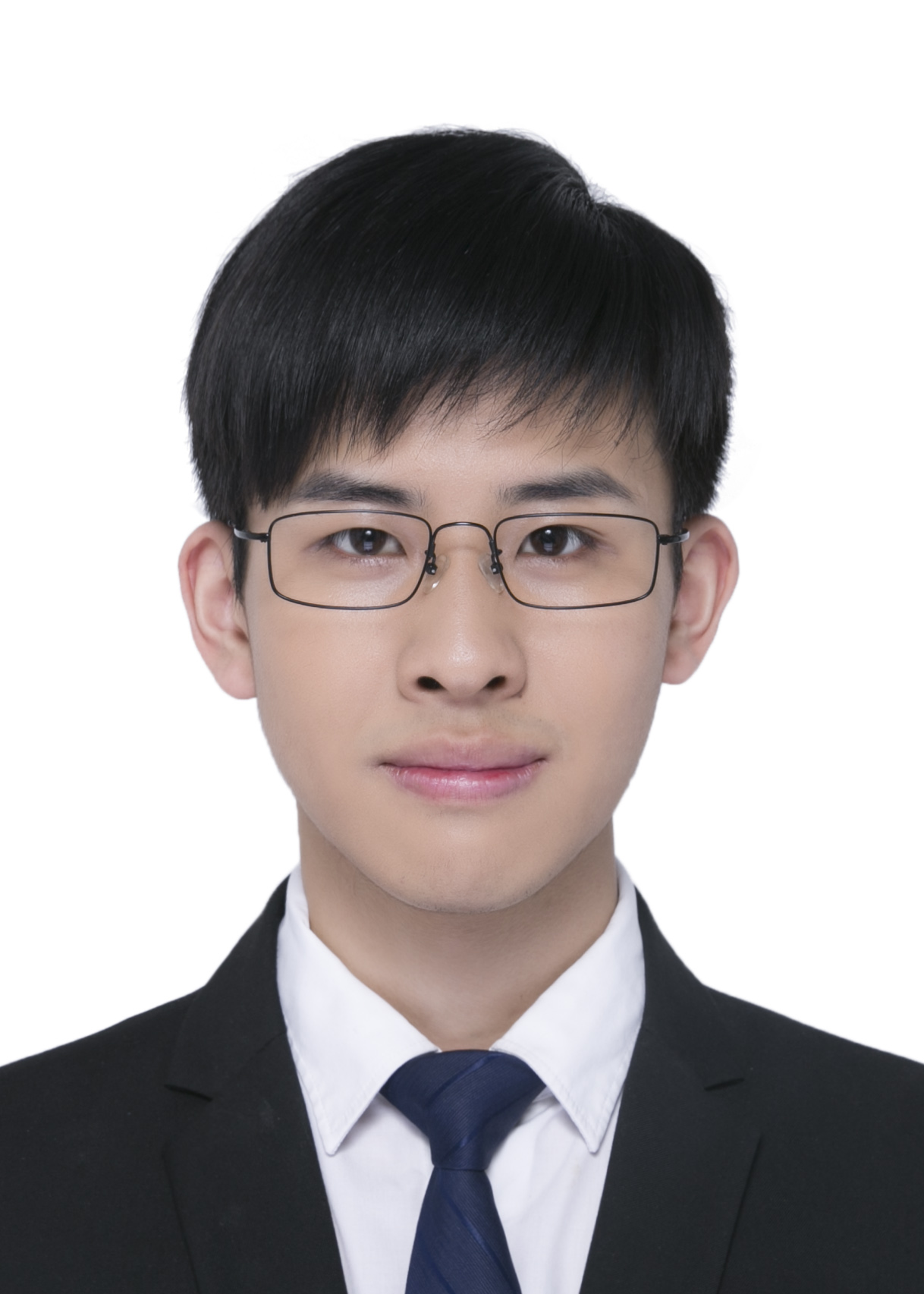}}]{Tianchen Zhao} received his B.S. degree in electronic engineering from Beihang University in 2020. He is currently pursuing his Ms. degree at EE Department, Beihang University. His research interest mainly focuses on efficient deep learning algorithm design.
\end{IEEEbiography}

\vspace{-1.0cm}
\begin{IEEEbiography}[{\includegraphics[width=1in,height=1.25in,clip,keepaspectratio]{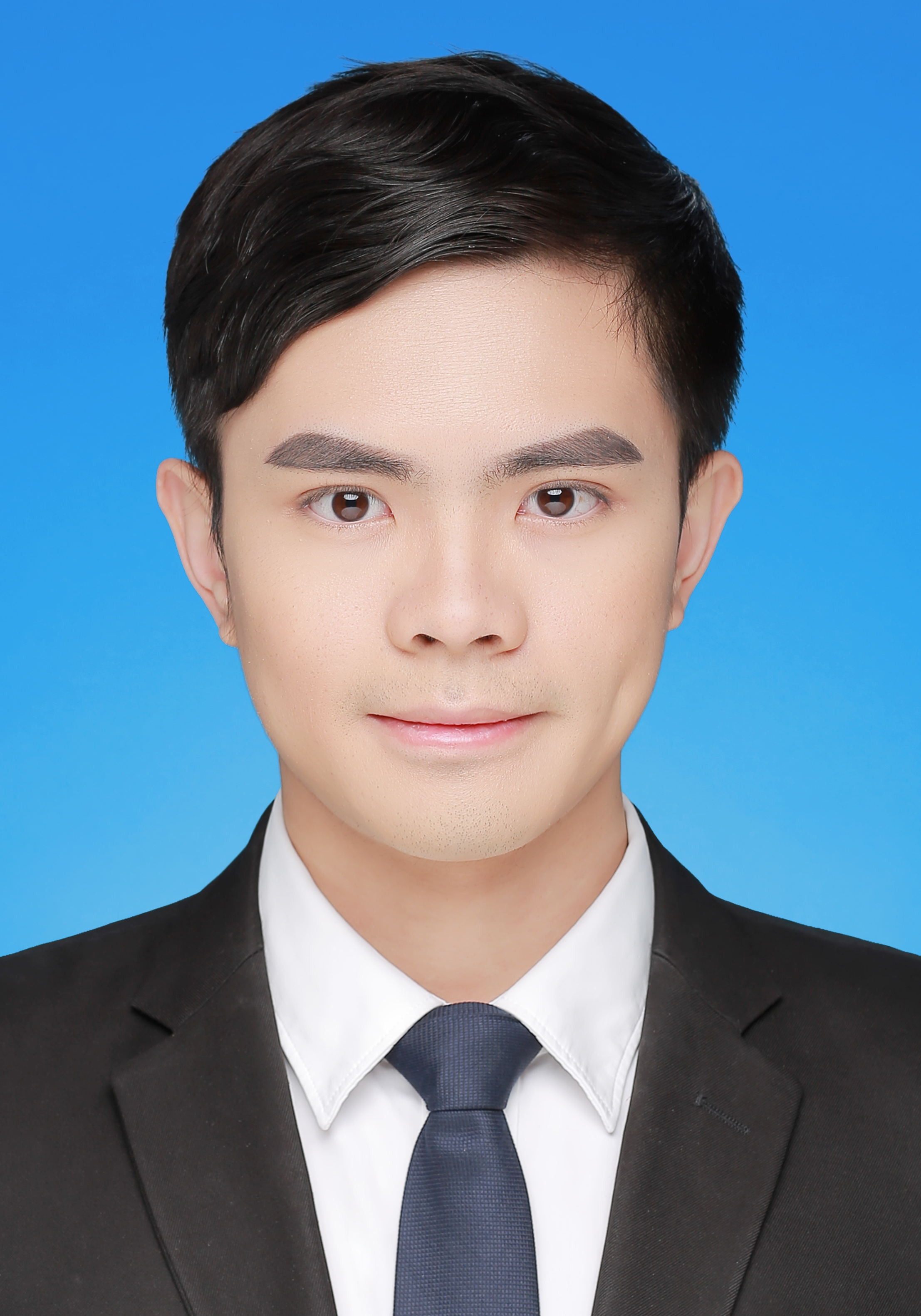}}]{Shulin Zeng}  received his B.S. degree in electronic engineering department of Tsinghua University, Beijing, China, in 2014. He is currently pursing his Ph.D. degree in electronic engineering department of Tsinghua University. His research mainly focuses on software-hardware co-design for deep learning and virtualization in the cloud.
\end{IEEEbiography}

\begin{IEEEbiography}[{\includegraphics[width=1.2in,height=1.4in,clip,keepaspectratio]{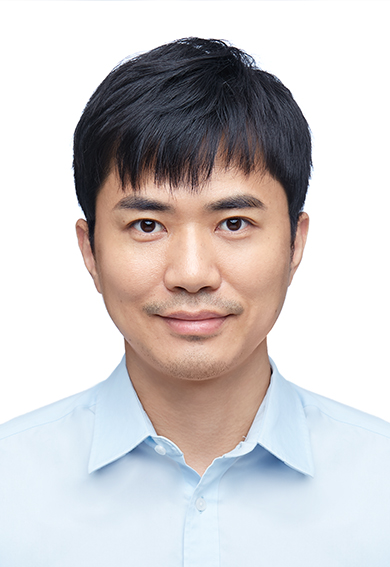}}]{Yu Wang}
(S'05-M'07-SM'14) received the B.S. and Ph.D. (with honor) degrees from Tsinghua University, Beijing, in 2002 and 2007. He is currently a tenured professor with the Department of Electronic Engineering, Tsinghua University. His research interests include brain inspired computing, application specific hardware computing, parallel circuit analysis, and power/reliability aware system design methodology. He has authored and coauthored more than 200 papers in refereed journals and conferences. He has received Best Paper Award in ASPDAC 2019, FPGA 2017, NVMSA 2017, ISVLSI 2012, and Best Poster Award in HEART 2012 with 9 Best Paper Nominations (DATE18, DAC17, ASPDAC16, ASPDAC14, ASPDAC12, 2 in ASPDAC10, ISLPED09, CODES09). He is a recipient of DAC under 40 innovator award (2018), IBM X10 Faculty Award (2010). He served as TPC chair for ICFPT 2019 and 2011, ISVLSI2018, finance chair of ISLPED 2012-2016, track chair for DATE 2017-2019 and GLSVLSI 2018, and served as program committee member for leading conferences in these areas, including top EDA conferences such as DAC, DATE, ICCAD, ASP-DAC, and top FPGA conferences such as FPGA and FPT. Currently, he serves as co-editor-in-chief of the ACM SIGDA E-Newsletter, associate editor of the IEEE Transactions on Computer-Aided Design of Integrated Circuits and Systems,the IEEE Transactions on Circuits and Systems for Video Technology, the Journal of Circuits, Systems, and Computers,and Special Issue editor of the Microelectronics Journal. He is now with ACM Distinguished Speaker Program. 
\end{IEEEbiography}

\begin{IEEEbiography}[{\includegraphics[width=1in,height=1.25in,clip,keepaspectratio]{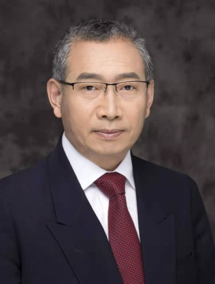}}]{Huazhong Yang}
(M'97-SM'00-F'20) received B.S. degree in microelectronics in 1989, M.S. and Ph.D. degree in electronic engineering in 1993 and 1998, respectively, all from Tsinghua University, Beijing. In 1993, he joined the Department of Electronic Engineering, Tsinghua University, Beijing, where he has been a Professor since 1998. Prof. Yang was awarded the Distinguished Young Researcher by NSFC in 2000, Cheung Kong Scholar by the Chinese Ministry of Education (CME) in 2012, science and technology award first prize by China Highway and Transportation Society in 2016, and technological invention award first prize by CME in 2019. He has been in charge of several projects, including projects sponsored by the national science and technology major project, 863 program, NSFC, and several international research projects. Prof. Yang has authored and co-authored over 500 technical papers, 7 books, and over 180 granted Chinese patents. His current research interests include wireless sensor networks, data converters, energy-harvesting circuits, nonvolatile processors, and brain inspired computing. He has also served as the chair of Northern China ACM SIGDA Chapter science 2014, general co-chair of ASPDAC’20, navigating committee member of AsianHOST’18, and TPC member for ASP-DAC’05, APCCAS’06, ICCCAS’07, ASQED’09, and ICGCS’10.
\end{IEEEbiography}


%\vfill


\end{document}